\documentclass[12pt,journal,compsoc,onecolumn]{IEEEtran}

\usepackage{amsmath,graphicx,cleveref,caption,subcaption,dsfont,multirow,multicol,amssymb,amsfonts,mathtools,pifont}
\usepackage[noadjust]{cite}
\usepackage[ruled,vlined]{algorithm2e}

\newtheorem{proposition}{Proposition}
\newtheorem{Proposition}{Proposition}

\usepackage{slashbox}
\usepackage{pifont}
\usepackage{xcolor}

\newcommand{\cmark}{\ding{51}}
\newcommand{\xmark}{\ding{55}}
\newcommand\bigSigma{\mbox{\large {\bf $\Sigma$}}}

\def\S{{\cal S}}

\title{FFNB: Forgetting-Free Neural Blocks for Deep  Continual Visual Learning}

\author{Hichem Sahbi$^1$ \ \ \ \   \ \ \ Haoming Zhan$^{1,2}$  \\ $^1$Sorbonne University, CNRS, LIP6
F-75005, Paris, France \\ $^2$XXII Group, Paris, France}

\begin{document}
\maketitle
\begin{abstract}
  Deep neural networks (DNNs) have recently achieved a great success in computer vision and several related fields. Despite such progress, current neural architectures still suffer from catastrophic interference (a.k.a. forgetting)  which obstructs DNNs to learn continually.  While several state-of-the-art methods have been proposed to mitigate forgetting, these existing solutions are either highly rigid (as regularization) or time/memory demanding (as replay). An intermediate class of methods, based on dynamic networks, has been proposed in the literature and provides a reasonable balance between task memorization and computational footprint. \\ In this paper, we devise a dynamic network architecture for continual learning based on a novel forgetting-free neural block (FFNB). Training FFNB features on new tasks is achieved using a novel procedure that constrains the underlying parameters in the null-space of the previous tasks, while training classifier parameters equates to Fisher discriminant analysis. The latter provides an effective incremental process which is also optimal from a Bayesian perspective. The trained features and classifiers are further enhanced using an incremental ``end-to-end'' fine-tuning. Extensive experiments, conducted on different challenging classification problems, show the high effectiveness of the proposed method. \\ 
  
\end{abstract}

\section{Introduction}
\label{sec:intro}\label{sec:intro}
Deep learning is currently witnessing a major success in different computer vision tasks including image and video classification \cite{jiu2016deep}. The purpose of deep learning is to train convolutional or recurrent  neural networks that map raw data into suitable representations prior to their classification \cite{mazari2019mlgcn,mazari2019deep}. However, the success of these networks is highly dependent on the availability of large collections of labeled training data that capture the distribution of the learned categories. In many practical scenarios, mainly those involving streams of data, large collections {\it covering the inherent variability of the learned categories} are neither available nor can be  holistically  processed. Hence,  training  deep networks should be achieved as a part of a {\it lifelong} process, a.k.a. continual or incremental learning~\cite{thrun1995lifelong, thrun1996learning, ring1994continual,sahbi2008manifold}. \\ 
\indent The traditional mainstream design of deep networks is based on  back propagation and stochastic gradient descent.  The latter collects gradients  through mini-batches and updates network parameters in order to learn different categories. However, in lifelong learning, tasks involve only parts of data/categories, and {this potentially leads to catastrophic forgetting (CF) defined as the inability of  a learning model to ``memorize''  previous tasks when handling new ones}. In cognitive science, CF is considered as  an extreme case of the stability-plasticity dilemma \cite{carpenter1987massively, mermillod2013stability} where excessive plasticity causes an easy fit to new knowledge and less on previous ones.  This is also related to  concept (or distribution) drift  \cite{gepperth2016incremental} that may happen when a learning model  keeps ingesting data~\cite{ioffe2015batch, santurkar2018does}.\\ 
\indent Whereas in most of the learning models (especially shallow ones~\cite{sahbi2008particular,dutta2017high,oliveau2017learning,wang2014bags,wang2014nonlinear}), CF could be overcome, its handling in deep networks is still a major challenge and existing solutions can only mitigate its effect. Indeed, CF results from the high non-linearity and entanglement of gradients when achieving back-propagation in deep networks (in contrast to shallow ones).  Existing  straightforward solutions  bypass this effect by storing huge collections of data  and {\it replaying} the learning process using all these collections; {whereas {\it replay} is highly effective, it is known to be time and memory demanding and may result into resource saturation even on sophisticated hardware devices}. Other solutions, with less time and memory footprint (e.g., regularization) can only mitigate the effect of CF.  Another category of methods, based on dynamic networks provides  a suitable balance between resource consumption and task memorization, and gathers  the advantage of the two aforementioned categories of methods (namely replay and regularization) while discarding their inconveniences  at some extent. Our proposed solution, in this work, is also built upon   dynamic networks and allows mitigating CF with a reasonable growth in the number of training parameters. 
\section{Related work}\label{related}
Early work in continual learning mitigates CF by constraining model parameters to keep previous knowledge while learning new tasks. These techniques include regularization, replay and dynamic networks. Elastic weight consolidation \cite{kirkpatrick2017overcoming} is one of the early regularization methods based on the Fisher information (see also \cite{liu2018rotate,ritter2018online}). Other criteria, including synaptic intelligence~\cite{zenke2017continual}, regularize parameters according to their impact on the training loss using gating mechanisms~\cite{serra2018overcoming} and weight pruning~\cite{mallya2018packnet}. Knowledge distillation~\cite{hinton2015distilling} has also been investigated to build cumulative networks that merge previous tasks with current ones~\cite{li2017learning}. These methods include  Incremental Moment Matching \cite{lee2017overcoming} which merges networks by minimizing a weighted Kullback–Leibler divergence and Learning without Memorizing \cite{dhar2019learning} that relies on attention mechanisms \cite{selvaraju2017grad}. Gradient episodic memory~\cite{lopez2017gradient} relies on a  memory budget but proceeds differently by regularizing and projecting the gradient of the current task onto the gradients of the previous ones. A variant in~\cite{chaudhry2018efficient}  extends further this regularization by averaging gradients through all the visited tasks while the method in \cite{castro2018end} combines limited representative memory with distillation. Other models seek to leverage extra knowledge including unlabeled data~\cite{zhang2019class,lee2019overcoming} (which are independent from the targeted tasks) or biases (due to imbalanced distributions) between previous and current tasks to further enhance generalization~\cite{wu2019large,hou2019learning}. \\  
\indent The second category of methods (namely replay) consists in leveraging original or generated pseudo data as exemplars for continual learning. ICaRL \cite{rebuffi2017icarl} is one of these methods which extracts image exemplars for each observed class depending on a predefined  memory budget. Pseudo-replay models \cite{kemker2017fearnet} including deep generative networks~\cite{shin2017continual,kamra2017deep,gulrajani2017improved,kingma2013auto,rostami2019generative} have also been investigated in the literature as alternative solutions that prevent storing exemplars. The particular method in \cite{li2019incremental} combines an explicit and an implicit memory where the former captures features from observed tasks and the latter corresponds to a discriminator and a generator similarly to deep generative replay. Other continual learning approaches employ coresets \cite{nguyen2017variational} to characterize the key information from different tasks. \\
\indent Closely related to our contribution, dynamic networks proceed by adapting the topology of the trained architectures either at a macroscopic or microscopic level ~\cite{yan2021dynamically}. Macroscopically, progressive networks \cite{rusu2016progressive} define parallel cascaded architectures where each sub-network characterizes a specific task. Each layer propagates its output {\it not only} in the same sub-network but also through all the sub-networks of the subsequent tasks. PathNet \cite{fernando2017pathnet} extends the topology of progressive networks, using evolutionary algorithms, to learn new connections between previous and current tasks. Random path selection networks \cite{rajasegaran2019random} push this concept further by learning potential skip-connections among parallel sub-networks using random search. Microscopically, existing methods dynamically expand networks using thresholds on loss functions over new tasks and retrain the selected weights to prevent semantic drift \cite{yoon2017lifelong}. Reinforced continual learning \cite{xu2018reinforced} employs a controller to define a strategy that expands the architecture of a given network while the learn-to-grow model \cite{li2019learn} relies on neural architecture search~\cite{zoph2018learning} to define optimal architectures on new tasks. Other models~\cite{draelos2017neurogenesis}, inspired by the process of adult neurogenesis in the hippocampus, combine architecture expansion with pseudo-rehearsal using auto-encoders. Our contribution in this paper proceeds differently compared to the aforementioned related work: tasks are incrementally handled by learning the parameters of a particular sub-network (referred to as FFNB) {\it in the null-space} of the previous tasks leading to {\it stable representations} on previously visited categories and discriminating representations {\it both} on previous and current categories. A bound is also provided that models the loss due to CF; this bound vanishes under particular settings of the null-space,  activations and weight decay regularization. All these statements are corroborated through extensive experiments on different classification problems.

\def\phii{{\psi}}
\def\phiip{{\hat{\psi}}}
\def\N{{\cal N}} 
\def\N{{\cal N}} 
\def\X{{\cal X}} 
\def\Y{{\cal Y}}  
\def\T{{\cal T}}  
 \def\W{{\bf W}} 
 \def\P{{\Phi}} 
 \def\XX{{\bf X}} 
 \def\alphaa{{\bf \alpha}} 
\def\YY{{\bf Y}}  
\def\I{{\bf I}}  
 \def\Sigmaa{{\bf \bigSigma}} 
 \def\S{{\bf S}} 
 \def\A{{\bf  A}}
\def\AA{{\cal A}}
 \def\B{{\bf  B}} 
\def\t{{\cal P}}
\def\r{{\cal P}}
\def\C{{\bf C}} 
\section{Problem formulation}\label{approach1}

Considering $\X$ as the union of input data (images, etc.) and $\Y$ their class labels drawn from an existing but unknown probability distribution $P(X,Y)$. The general goal is to train a  network $f: \X \rightarrow \Y $ that assigns a label $f(X)$ to a given sample $X$ while minimizing a generalization risk $R(f)=P(f(X)\neq Y)$.  The design of $f$ is usually achieved by minimizing an objective function (or loss) on a {\it fixed}  set $\T=\{(x_i,y_i)\}_i$ including all the training data and their labels; this scheme is known as multi-task learning~\cite{jiu2016laplacian,jiu2019deep}.  In contrast, we consider in this work a different setting which learns $f$ incrementally, i.e., only a subset of $\T$ (denoted $\T_t$) is available at a given cycle $t$ (see for instance algorithm~\ref{alg1} in supplementary material). In what follows $\T_t$  will be referred to as {\it task}.\\ 
\indent Let  $\{\T_1,\dots,\T_T\}$ be a collection of tasks; this set is not necessarily a partition of $\T$, and each $\T_t$ may include one or multiple classes. Learning $f$ incrementally may lead to CF; {the latter is defined as the inability of $f$ to remember (or correctly classify) previously seen tasks either due to a {\it distribution-shift} (i.e., when tasks correspond to the same classes but drawn from different distributions) or to a {\it class-shift} (i.e., when tasks correspond to disjoint classes)}. We consider only the {\it latter} while the {\it former} (closely related to domain adaptation) is out of the scope of this paper. 

\subsection{Dynamic networks and catastrophic forgetting}

Without a loss of generality, we consider $T$ as an overestimated (maximum) number of tasks visited during a continual learning process. We also consider $f$ as a convolutional network whose fully connected (FC) layers are dynamically updated. This FC sub-network of $f$, also referred to as FFNB (see Table.~\ref{fig111111a}), corresponds to feature maps and classification layers whose widths $\{d_\ell\}_\ell$  (or dimensionalities) are dynamically expanded as tasks evolve. This dimensionality expansion makes data, belonging to the current and the previous tasks, increasingly separable. Let  $\XX_t \in \mathbb{R}^{d_0 \times n_t}$ denote the data matrix of a given task $t$ {(with $n_t=|\T_t|$)}, the maps of these FC layers are recursively defined as $\phii_{\ell}(\XX_t) = g\big(\W_{\ell} \  \phii_{\ell-1}(\XX_t)\big)$ with  $\phii_{\ell}(.) \in \mathbb{R}^{d_\ell}$, $\ell \in \{1,\dots,L\}$ and $\phii_{0}(\XX_t)=\XX_t$. Here $\W_{\ell}  \in \mathbb{R}^{d_\ell \times d_{\ell-1}}$ is a matrix of training parameters and $g$ a nonlinear activation. Considering this dynamic network, a band of parameters is assigned to each new task and trained ``end-to-end'' by back-propagating the gradient of a {\it task-wise} loss. These parameters are afterwards updated while those assigned to the previous tasks remain unchanged; see again algorithm~\ref{alg1} in supp material.\\
\indent As data belonging to the previous tasks are dismissed, this straightforward ``end-to-end'' learning of the current task's parameters relies only on data in $\T_t$, and thereby the underlying classifier may suffer from insufficient generalization. Moreover, as no updates are allowed on the parameters of  $\T_1,\dots,\T_{t-1}$, it follows that {\it neither} generalization {\it nor}  CF are appropriately handled on the previous tasks; {indeed, as the parameters $\{\W_{\ell,t}\}_\ell$ of a given task $t$ evolve, there is no guarantee that the outputs $\{\phii_\ell\}_\ell$ remain unchanged on data of the previous tasks,  leading to changes in the underlying classification output $\phii_{L+1}$ and thereby to CF (see later experiments)}. One may consider the statistics of the previous tasks (e.g., means and covariances of the data across different FC layer maps) and discriminatively learn the parameters associated  to the current task (see later section~\ref{FDAa}). Nevertheless, prior to this step, one should be cautious in the way learning is achieved with those statistics, as the latter should remain stable as tasks evolve. Indeed, even when those statistics are available, one cannot update the {\it parameters} of the previous tasks as the {\it latter} disrupt in turn those statistics and no data are available in order to re-estimate them on the previous tasks. Hence, before making updates on the network parameters, feature maps should be stabilized  on the previous tasks as introduced subsequently.

\subsection{Proposed method} 
 We introduce in this section an alternative solution which sill relies on dynamic networks but considers different FC layers and training procedure. {Our framework incrementally learns the parameters $\{\W_{\ell,t}\}_\ell$ of the current task $\T_t$ in the null-space of the previous tasks  $\T_1,\dots, \T_{t-1}$ while maintaining the dynamic outputs  $\{\phii_{\ell}\}_\ell$ of all the FC/FFNB layers almost unchanged (or at least stable) on $\T_1,\dots, \T_{t-1}$}. This approach, as described subsequently, learns {\it new tasks} incrementally and mitigates CF on {\it  previous ones} while maintaining high generalization {\it on both}.   
\subsubsection{FFNB features} 
\noindent  Learning the parameters of the current task should guarantee:  (i) the {\it consistency} of the network predictions w.r.t. the underlying ground-truth, and (ii) the {\it stability} of the feature maps of the FC layers on the previous tasks. The first constraint is implemented by minimizing  a hinge loss criterion while the second one is  guaranteed by constraining the parameters of a new task to lie  in the null-space $\N_S(\phii_\ell(\XX_\t))$ of previous tasks data; in this notation,  $\t=\{1,\dots,t-1\}$ and $\XX_\t$ refers to  the matrix of data in  $\T_1,\dots,\T_{t-1}$. As shown subsequently, stability is implemented by learning the parameters of a new task in a residual subspace spanned by the axes of principal component analysis (PCA)\footnote{\scriptsize with the smallest statistical variance.} applied to $\phii_\ell(\XX_\t)$. \\ 
\indent  Let $\P^{\ell,t}$ be the matrix of eigenvectors (principal axes of PCA) associated to data in $\phii_\ell( \XX_\t)$; in what follows, unless stated otherwise, we write  $\P^{\ell,t}$ simply as  $\P$. Assuming these data centered, the principal axes are obtained by diagonalizing a covariance matrix  incrementally defined as  $\sum_{r=1}^{t-2} \phii_\ell(\XX_{r}) \phii_\ell(\XX_{r})^\top +  \phii_\ell(\XX_{t-1}) \phii_\ell(\XX_{t-1})^\top$. The eigenvectors $\{\P_{d}\}_d$ in $\P$ constitute an orthonormal basis sorted following a decreasing order of the underlying eigenvalues. Let $p$ be the smallest number of dimensions which concentrate most of the statistical variance. The vector of parameters associated to the current task $t$ in  $\N_S(\phii_\ell(\XX_\t))$ is 
\begin{equation}\label{eq1}
  \W_{\ell,t} := \sum_{d=p+1}^{d_{\ell-1}} \alphaa_{\ell,t}^d \P_d^\top,
\end{equation} 

\noindent and training the latter equates to optimizing $\alpha_{\ell,t}=(\alphaa_{\ell,t}^{p+1},\dots,\alphaa_{\ell,t}^{d_{\ell-1}})^\top$. Let $E$ denote a loss function associated to our classification task;  considering the aforementioned reparametrization of $\W_{\ell,t}$, the gradient of the loss is now updated using the chain rule  as  $\frac{\partial E}{\partial \alphaa_{\ell,t}} = \frac{\partial E}{\partial \W_{\ell,t}} \ \frac{\partial \W_{\ell,t}}{\partial  \alphaa_{\ell,t}}$ being $\frac{\partial E}{\partial  \W_{\ell,t}}$ the original gradient  obtained using    back-propagation as provided with  standard deep learning frameworks (including PyTorch and TensorFlow) and $\frac{\partial \W_{\ell,t}}{\partial  \alphaa_{\ell,t}}$ being a Jacobian matrix. The latter --- set with the $(d_{\ell-1}-p)$  residual PCA eigenvectors  ---  is used to maintain $\W_{\ell,t}$ in the feasible set, i.e., $\N_S(\phii_\ell(\XX_\t))$. Considering this update scheme, the following proposition shows the consistency of the training process when handling CF.\\

\begin{proposition}
  Let $g: \mathbb{R} \rightarrow \mathbb{R}$ be a $L$-Lipschitz continuous activation (with $L\leq 1$). Any $\eta$-step update of  $\W_{\ell,t}$ in $\N_S(\phii_\ell(\XX_\t))$ using (\ref{eq1}) satisfies $\forall r \in \t$
  \begin{equation}\label{eq3} \small
    \begin{array}{lll}
      \big\|\phii_\ell^{\eta-1}({\XX}_r)- \phii_\ell^0({\XX_r})  \big\|_F^2  \leq B \\
     \\

  \textrm{with}  \ \ \  \ \ \  \ \ \       B = \displaystyle \sum_{\tau=1}^{\eta-1}  \sum_{k=0}^{\ell-1}   \big(\big\| \alpha_{\ell-k,t}^\tau\big\|_F^2 . \big\| \beta_{\ell-k-1,r}^{\tau}   \big\|_F^2 +  \big\| {\alpha}_{\ell-k,t}^{\tau-1}\big\|_F^2  . \big\| {\beta}_{\ell-k-1,r}^{\tau-1} \big\|_F^2 \big). \prod_{k'=0}^{k-1}  \big\|\W_{\ell-k',\r}^{\tau} \big\|_F^2,
\end{array}
       \end{equation} 
being  $\phii_\ell^{0}({\XX}_r)$  (resp. $\phii_\ell^{\eta-1}({\XX}_r)$) the map before the start (resp. the end) of the iterative update (gradient descent on current task $\T_t$), $\beta_{\ell,r}^\tau$ the projection of $\phii_\ell^{\tau}({\XX}_r)$ onto $\N_S(\phii_\ell(\XX_\t))$ at any iteration $\tau$, $\{\W_{\ell,r}^\tau\}_\ell$ the network parameters at $\tau$, and $\|.\|_F$ the  Frobenius norm.
\end{proposition} 
 \def\tr{{\bf Tr}}
 \noindent Details of the proof are omitted and can be found in the supplementary material. More importantly, the bound in Eq.~\ref{eq3} suggests that FFNB layers endowed with $L$-Lipschitzian activations (e.g., ReLU)  and low statistical variance in $\N_S(\phii_\ell(\XX_\t))$ make CF contained. Eq.~\ref{eq3} also suggests that one may use weight decay (on $\{\alpha_{\ell,t}\}$)  to regularize the parameters $\{\W_{\ell,t}\}_{t,\ell}$ leading to a tighter bound $B$, and again contained CF. Note  that Eq.~\ref{eq3} is an increasing function of $\ell$, so shallow FC layers suffer less from CF compared to deeper ones. However, controlling the norm of  $\{\W_{\ell,t}\}_{t,\ell}$ (and $p$ in Eq.~\ref{eq1}) effectively mitigates the effect of CF (due to the depth) and maintains generalization. As a result, all the statistics (mean and covariance matrices) used to estimate the eigenvectors of PCA and also to update the classifier parameters (in section~\ref{FDAa}) remain stable. In short, Eq.~\ref{eq1} provides an effective way to ``memorize'' the previous tasks\footnote{\scriptsize Note that Eq.~\ref{eq1} leads to almost orthogonal parameters  through successive tasks (with shared residual components), and this provides an effective way to leverage both ``shared multi-task'' and ``complementary'' informations despite learning incrementally.}. 

  \subsubsection{Initialization}
We introduce a suitable initialization of the feature map parameters $\{{\alpha_{\ell,t}}\}_{\ell,t}$  which turns out to be effective during optimization (fine-tuning). We cast the problem of setting  $\{{\alpha_{\ell,t}}\}_{\ell,t}$  (or equivalently $\{\W_{\ell,t}\}_{\ell,t}$; see Eq.~\ref{eq1}) as a solution of the following regression problem

\begin{equation}\label{eq5} 
  \min_{\alpha_{\ell,t}}  \frac{\gamma}{2} \big\|\alpha_{\ell,t}\big\|_F^2  \ + \   \frac{1}{2}  \big\| \C_{\ell,t} - \alpha_{\ell,t}^\top \ \P^\top \  \psi_\ell(\XX_{t}) \big\|_F^2,
  \end{equation} 
here $\C_{\ell,t} \in \mathbb{R}^{d_{\ell} \times n}$ {(with $n=\sum_{r\leq t} |\T_{r}|$)} is a predefined coding matrix whose entries are set to $+1$ iff data belong to current task $t$ and $0$ otherwise. One may show that optimality conditions (related to the gradient of Eq.~\ref{eq5}) lead to the following solution
\begin{equation}\label{eq6} 
\alpha_{\ell,t}=\big( \gamma  \I +\P \  \psi_\ell(\XX_t) \  \psi_\ell(\XX_{t})^\top \  \P^\top \big)^{-1}  \P \ \psi_\ell(\XX_{t}) \  \C_{\ell,t}^\top.
\end{equation} 
\noindent As the setting of $\alpha_{\ell,t}$ relies only on current (mono) task data, it is clearly sub-optimal and may affect the discrimination power of the learned feature maps. In contrast to the above initialization, we consider another (more effective multi-task) setting using all $\{\T_r\}_{r\in \AA}$ (with $\AA=\t\cup \{t\}$), and without storing all the underlying data. Similarly to Eq.~(\ref{eq6}), we derive 
\begin{equation}\label{eq7}
  \alpha_{\ell,t}=\big( \gamma  \I +\P \ \sum_{r \in \AA} [\psi_\ell(\XX_r) \ \psi_\ell(\XX_r)^\top] \ \P^\top \big)^{-1} \ \P \ \sum_{r \in \AA} [\psi_\ell(\XX_r) \ \C_{\ell,r}^\top].
  \end{equation} 
{This equation can still be evaluated incrementally (without forgetting) while leveraging multiple tasks $\T_1,\dots,\T_t$, and without explicitly storing the whole data in $\{\psi_\ell(\XX_{r})\}_{r}$ and $\{\C_{\ell,r}\}_{r}$}. 

\subsubsection{FFNB classifiers}\label{FDAa}

The aforementioned scheme is applied  in order to learn the parameters of the feature maps while those of the classifiers are designed differently. The setting of these parameters is based on Fisher discriminant analysis (FDA) which has the advantage of being achievable incrementally by storing only the means and the covariance matrices associated to each task. Given current and previous tasks (resp. $\T_t$ and $\T_r$), FDA approaches the problem by modeling the separable FFNB features (as designed earlier) as gaussians with means and covariances  $(\mu_{t}^{L-1},\Sigmaa_{t}^{L-1})$, $(\mu_{r}^{L-1},\Sigmaa_{r}^{L-1})$ respectively.  Following this assumption, the Bayes optimal decision function corresponds to the log likelihood ratio test. One may show that its underlying separating hyperplane  $(\W_{L,(t,r)},b)$ maximizes the following objective function

  \begin{equation}
  \begin{array}{l}
\max_{ \W_{L,(t,r)}}  \frac{ \big(\W_{L,(t,r)}^\top \ \big (\mu_{t}^{L-1} - \mu_{r}^{L-1}\big)\big)^2}{ \W_{L,(t,r)}^\top \  \big(\Sigmaa_{t}^{L-1}+\Sigmaa_{r}^{L-1}\big) \  \W_{L,(t,r)}  }, 
    \end{array} 
  \end{equation} 
with its shrinkage estimator  solution being
\begin{equation}\label{eq111}
    \begin{array}{lll}
 \W_{L,(t,r)} &=& \big(\Sigmaa_{t}^{L-1} + \Sigmaa_{r}^{L-1} + \epsilon \I \big)^{-1} \big(\mu_{t}^{L-1}  - \mu_{r}^{L-1}\big) \\ 
 b &=&  -  \W_{L,(t,r)}^\top (\mu_{t}^{L-1} + \mu_{r}^{L-1}\big).
    \end{array} 
\end{equation} 
\noindent Under heteroscedasticity (i.e., $\Sigmaa_{t}^{L-1}\neq \Sigmaa_{r}^{L-1}$), these covariance matrices allow normalizing the scales of classes leading to better performances as shown later in experiments. Considering the pairwise class parameters $\{\W_{L,(t,r)}\}_{r \in \t}$,  one may incrementally derive the underlying pairwise classifiers as 

\begin{equation} 
  \phii_L^{(t,r)} (.) = \textrm{tanh}  \big(\W_{L,(t,r)} \ \phii_{L-1} (.)\big),
  \end{equation} 
and the output of the final classifier $ \phii_{L+1}^{t} (.)$ (see again Table.~\ref{fig111111a}) is obtained by pooling all the pairwise scores $\{\phii_L^{(t,r)} (.)\}_{t,r}$ through $r\in \t$ resulting into
\begin{equation} 
  \phii_{L+1}^{t}(.)=  \sum_{r\in \t} \phii_L^{(t,r)} (.).
\end{equation}
\noindent As shown later in experiments (see supp material), these incremental aggregated  ``one-vs-one'' classifiers outperform the usual ``one-vs-all'' softmax while mitigating CF. This also follows the learned separable FFNB features which make these classifiers highly effective.

\subsubsection{``End-to-end''  fine-tuning}\label{ft}
End-to-end fine-tuning of the whole network involves only the parameters of the feature map and the classification layers associated to the current task, as the other parameters cannot be updated without knowing explicitly the data. {Note that the convolutional layers are kept fixed: on the one hand, these layers capture low-level features, which are common to multiple tasks and can therefore be pre-trained offline. On the other hand, retraining the convolutional layers may disrupt the outputs of the feature maps and hence the classifiers.} Details of the whole ``end-to-end'' incremental learning are described in algorithm~\ref{alg2} in supp material.

\section{Experimental validation}
We evaluate the performance of our continual learning framework on the challenging task of action recognition, using the SBU and FPHA datasets \cite{Yun2012,Garcia2018}. SBU is an interaction dataset acquired (under relatively well controlled conditions) using the Microsoft Kinect sensor; it includes in total 282 moving skeleton sequences (performed by two interacting individuals) belonging to 8 categories. Each pair of interacting individuals corresponds to two 15 joint skeletons and each joint is encoded with a sequence of its 3D coordinates across video frames~\cite{mazari2019mlgcn}. The FPHA dataset includes 1175 skeletons belonging to 45 action categories which are performed by 6 different individuals in 3 scenarios. Action categories are highly variable with inter and intra subject variability including style, speed, scale and viewpoint. Each skeleton includes 21 hand joints and each joint is again encoded with a sequence of its 3D coordinates across video frames~\cite{mazari2019mlgcn}. In all these experiments, we use the same evaluation protocol as the one suggested in \cite{Yun2012,Garcia2018} (i.e., train-test split\footnote{\scriptsize excepting that training data belonging to different classes are visited incrementally.}) and we report the average accuracy over all the {\it visited} classes of actions\footnote{{\scriptsize Due to space limitation,  extra experiments and comparisons can be found in the supplementary material}.}.  
\subsection{Setting and performances}\label{baselines}  
The whole network architecture is composed of a {\it spatial graph convolutional} block similar to~\cite{wu2020comprehensive} appended to {\it FFNB}. The {\it former} includes an aggregation layer and a dot product layer while the  {\it latter} consists of our feature map and classification layers \cite{sahbi2021learning}.  During incremental learning, all the parameters are fixed excepting those of the current task (in FFNB) which are allowed to vary. For each task, we train the network parameters (FC layers) as described earlier for $250$ epochs per task  with a batch size equal to $50$, a momentum of $0.9$ and a learning rate (denoted as $\nu(t)$)   inversely proportional to the speed of change of the current task loss; when this speed increases (resp. decreases),   $\nu(t)$  decreases as $\nu(t) \leftarrow \nu(t-1) \times 0.99$ (resp. increases as $\nu(t) \leftarrow \nu(t-1) \slash 0.99$). All these experiments are run on a GeForce GTX 1070 GPU device (with 8 GB memory) and no data augmentation is achieved. \\
\begin{table}
  \begin{center}
\resizebox{1\textwidth}{!}{\begin{tabular}{c||cccccccc}
  \backslashbox{PCA dim}{ \hspace{1cm} Tasks (1 class / $\T_t$)}    & $\T_1(1)$ & $\T_2 (2) $ & $\T_3 (3)$ & $\T_4 (4)$ & $\T_5 (5)$ & $\T_6 (6)$ & $\T_7 (7)$ & $\T_8 (8)$  \\  
  \hline
  \hline
   $p=15$      & 100.00& 100.00& 96.42& 97.36& 95.34& 87.50&  84.48 & 81.53   \\
    $p=25$     & 100.00& 100.00& 100.00& 97.36& 95.34& 85.41& 82.75& 83.07    \\
   $p=35$      & 100.00& 100.00& 96.42 & 97.36& 95.34& 85.41& 82.75& 81.53    \\
   $p=45$      & 100.00& 100.00& 100.00& 100.00& 97.67& 89.58& 89.65& \bf84.61 \\
                             $p=50$      & 100.00& 100.00& 96.42 & 97.36 & 93.02&  85.41&  81.03&  73.84\\
                          $p=55$  &  100.00& 100.00& 100.00&  97.36&   90.69&  87.50&  75.86&  69.23 \\ 
                             \hline
Incremental (FFNB, algorithm 1)   &   100.00 & 50.00&  28.57&  26.31&  18.60&  16.66&  13.79&   12.30  \\
      Multi-task (all, upper bound) &            ---    &---    &---    &---    &---    &---    &---    & \bf90.76  \\
 \hline 
                        {Network size (\# of param)}    &  {2165}    & {2372}      &     {2619}  &   {2924}    &   {3305}    &   {3780}      &  {4367}     & {5084} \\                         
                             {Average training time per epoch (in seconds)}    &    {0.1326} & {0.1369}  &  {0.1411}  &  {0.1475} &  {0.1519}  &   {0.1651}  &   {0.1629}  &  {0.1737}

\end{tabular}}
\end{center}\caption{\scriptsize Impact of retained PCA dimension on the performances of incremental learning (the max nbr of dimensions is 60); for each column (task $\T_t$), performances are reported on the union of all the visited classes (i.e., $[1-t]$).}\label{table1} 
\end{table}
\begin{table}  
  \begin{center}
\resizebox{1\textwidth}{!}{\begin{tabular}{c||ccccccccc}
                             \backslashbox{PCA dim}{Tasks \\ (5 classes/$\T_t$)}    & $\T_1$ & $\T_2$ & $\T_3$ & $\T_4$ & $\T_5$ & $\T_6$ & $\T_7$ & $\T_8$ & $\T_9$ \\                                                       & $[1-5]$ & $[6-10]$ & $[11-15]$ & $[16-20]$ & $[21-25]$ & $[26-30]$ & $[31-35]$ & $[36-40]$ & $[41-45]$ \\

  \hline
                             \hline

$p=15$      &  65.07  &   57.14 &  55.44 &  57.52 &  58.51 &  56.07 &  59.42 &  62.10 &  62.26 \\ 
$p=30$      &  65.07  &   55.55 &  56.99 &  57.52  & 56.96 & 56.84 &  59.42  & 60.74  & 61.39 \\ 
$p=45$      &  68.25  &   57.14&   60.10&   58.30&   56.03&   55.03&   60.97&  62.30 &   62.60 \\ 
$p=60$      &  68.25  & 55.55 &  62.17 &  60.61 &  63.15 &  62.53  & 65.41 &  64.84  & 66.08 \\ 
$p=75$      &  68.25  & 58.73  & 64.76  & 63.70 &  62.84  & 61.24 &  64.96  & 66.01  & \bf67.47 \\ 
$p=90$      &  69.84 &  61.90  & 67.87  & 62.93 &  60.06  & 59.94 &  61.41 & 63.28 &  62.78 \\
                             $p=105$     &  65.07  & 62.69 & 68.39 & 63.32  & 58.51 &  58.65  & 60.97  & 62.50  & 62.43 \\
                             $p=120$     &  61.90 &  60.31 &  65.80  & 62.54 &  57.27 &  51.93 & 56.54 &55.27  & 56.86   \\                                   \hline
Incremental (FFNB, algo 1)   & 19.04  &  9.52  &  6.21  &  4.63   & 3.71 &  3.10  &  2.66  &  2.34  &  2.08 \\ 
      Multi-task (all, upper bound) &            ---    &---    &---    &---    &---    &---    &---    & --- & \bf84.17 \\
 \hline 
                             {Network size (\# of param)}    &  {5549}      &   {8784}    &  {14819}   &  {25904}   &  {44289}  &   {72224}     &  {111959}     &  {165744}  &  {235829} \\  
{Avg. training time per epoch (in s)}    & {2.0373}   &  {2.0241} & {2.0686} & {2.1463} &  {2.2393} &  {2.3318} &    {2.3992} &  {2.5082} &  {2.5755}
\end{tabular}}
\end{center}\caption{\scriptsize Impact of retained PCA dimension on the performances of incremental learning (the max nbr of dimensions is 168); for each column (task $\T_t$), performances are reported on the union of all the visited classes (i.e., $[1-5t]$).}\label{table12} 
\end{table} 
\noindent Tables.~\ref{table1} and~\ref{table12} show the behavior of our continual learning model w.r.t.  $p$ the number of dimensions kept in PCA. From these results, it becomes clear that 45 dimensions on SBU (75 on FPHA) capture most of the statistical variance of the previous tasks ($\gg 95\%$ in practice) and enough dimensions are hence reserved to the current task. These dimensions make the learned FFNB stable on the previous tasks while also being effective on the current one.  These tables also show  a comparison of incremental   and multi-task learning baselines (which learn the convolutional block and FFNB  ``end-to-end'' using the whole ambient space). Note that multi-task learning performances are available only at the final task as this baseline requires all the tasks. From these tables, it is clear that the multi-task baseline obtains the best performance, however, our proposed method reaches a high accuracy as well in spite of being incremental  while the second (incremental) baseline behaves almost as a random classifier.  Tables~\ref{table1}, \ref{table12} also show network size {and training time} as tasks evolve while tables~\ref{tableaa},~\ref{tablebb},~\ref{tablehh},~\ref{tableii} (in the supp material) show extra-tuning w.r.t. respectively the number of layers and {\it band-sizes}; the latter correspond to the {\it number of added neurons} per layer and per task.  
\subsection{Ablation study and comparison}\label{ablation}
We also study the impact of each component of our continual learning model on the performances  when taken separately and jointly.  From the results in tables~\ref{tablecc} and~\ref{tabledd}, the use of the null-space (in Eq.~\ref{eq1}) provides a significant gain in performances and  the impact of FDA covariance normalization (i.e., heteroscedasticity in Eq.~\ref{eq111}) is also globally positive. This results from the learned features, in  FFNB, which are designed to be {\it separable} without necessarily being {\it class-wise homogeneous}, and thereby their normalization  provides an extra gain  (as again shown through these performances)\footnote{\scriptsize It's worth noticing that our normalization is different from the usual batch-normalization, as the latter allows obtaining homogeneous features through neurons belonging to the same layer while our normalization makes features homogeneous through classes/tasks (see also empirical evidences in tables~\ref{table6aaaa} and~\ref{table6bbbb} of the supp material).}.  We also observe the positive impact of multi-task initialization w.r.t. mono-task and random initializations (Eq.~\ref{eq7} vs.~\ref{eq6}). Finally,  tables~\ref{table6} and~\ref{table62} (in supp material) show the impact of network pre-training and ``end-to-end'' fine-tuning on the performances (see again section~\ref{ft}). As observed, the best performances are obtained when both pre-training and fine-tuning are used.
\begin{table}
  \begin{center}
    \resizebox{0.9\textwidth}{!}{
    \begin{tabular}{ccc|cccccccc}
      Null-space& Heteroscedasticity & Init  &   $\T_1(1)$ & $\T_2 (2) $ & $\T_3 (3)$ & $\T_4 (4)$ & $\T_5 (5)$ & $\T_6 (6)$ & $\T_7 (7)$ & $\T_8 (8)$  \\
      \hline
      \hline 
   \xmark   & \xmark & rand & 100.00 & 85.00 &67.85& 55.26& 44.18& 37.50& 34.48& 29.23   \\
   \xmark   & \xmark & mono  &100.00 & 90.00 & 53.57 & 36.84& 27.90& 22.91& 25.86& 26.15   \\
   \xmark   & \xmark & multi & 100.00 &90.00& 71.42 &68.42 &51.16& 45.83& 46.55& 43.07   \\
      \hline
  \xmark   & \cmark & rand & 100.00& 100.00 &96.42& 86.84& 51.16& 54.16& 63.79& 56.92  \\
   \xmark   & \cmark &  mono & 100.00& 50.00 &53.57& 26.31& 11.62& 12.50& 20.68& 10.76     \\
   \xmark   & \cmark & multi & 100.00& 100.00& 100.00& 47.36& 53.48& 39.58& 44.82& 44.61 \\
      \hline
      \hline
   \cmark   & \xmark  & rand &100.00 &90.00& 53.57& 44.73& 37.20& 31.25& 31.03& 30.76    \\
   \cmark   & \xmark & mono & 100.00 &95.00 &92.85 &94.73& 81.39& 68.75& 60.34& 50.76     \\
   \cmark   & \xmark & multi &100.00& 100.00& 100.00& 97.36& 90.69& 87.50& 82.75& 75.38  \\
   \hline 
   \cmark   & \cmark & rand &  100.00 &50.00& 57.14& 50.00& 25.58& 29.16& 34.48& 30.76     \\
   \cmark   & \cmark & mono &100.00& 100.00& 96.42& 94.73& 72.09& 54.16& 50.00 &55.38   \\
   \cmark   & \cmark & multi & 100.00 &100.00& 100.00& \bf100.00& \bf97.67& \bf89.58& \bf89.65& \bf84.61 
\end{tabular}}
\end{center}\caption{\scriptsize Ablation study (with pretraining and fine-tuning, here $p=45$); for each column (task $\T_t$), performances are reported on the union of all the visited classes (i.e., $[1-t]$).}   \label{tablecc}
\end{table}
\begin{table}
  \begin{center}
    \resizebox{1\textwidth}{!}{
    \begin{tabular}{ccc|ccccccccc}
      Null-space& Heteroscedast. & Multi  &   $\T_1$ & $\T_2$ & $\T_3$ & $\T_4$ & $\T_5$ & $\T_6$ & $\T_7$ & $\T_8$ & $\T_9$ \\
                &                    &        &  $[1-5]$ & $[6-10]$ & $[11-15]$ & $[16-20]$ & $[21-25]$ & $[26-30]$ & $[31-35]$ & $[36-40]$ & $[41-45]$ \\
      \hline
      \hline
       \xmark   & \xmark & \xmark &  19.04  & 11.11  & 9.32 &  5.79  &  7.43 &   5.68  &  6.43  &  5.85  &  6.08 \\
   \xmark   & \xmark & \cmark &    71.42 &  65.07 &  59.06 & 59.84 &  62.84 & 60.72 &  59.86 &  58.20 &  56.52 \\
      \hline
   \xmark   & \cmark & \xmark &  19.04 &  11.11 &  6.21  &  5.40  & 4.02 &  3.10  &  3.32 &  2.73  &  2.08  \\
   \xmark   & \cmark & \cmark &  69.84 &  56.34 &  54.92 &  54.44  & 57.89  & 60.46 &  64.07 & 64.84  & 65.39  \\
      \hline
      \hline
   \cmark   & \xmark & \xmark &    77.77 &  67.46  & 66.83  & 63.32 &  60.99 &  58.39 &  58.98  & 57.42 &  56.00 \\
   \cmark   & \xmark & \cmark & 76.19 &  67.46 &  64.76 &  62.93 &  62.84 &  61.24  & 59.42 &  58.00 &  54.95  \\
   \hline 
   \cmark   & \cmark & \xmark &   71.42  & 71.42 &  62.69  & 54.05 &  54.48 & 47.80 & 47.45  & 45.50 &  45.56 \\
   \cmark   & \cmark & \cmark &   68.25  & 58.73  & 64.76  & 63.70 &  62.84  & 61.24 &  64.96  & 66.01  & \bf67.47          
\end{tabular}}
\end{center}\caption{\scriptsize Ablation study (with pretraining and fine-tuning, here $p=75$); for each column (task $\T_t$), performances are reported on the union of all the visited classes (i.e., $[1-5t]$).}   \label{tabledd}
\end{table}
\subsection{{Extra experiments: CIFAR100}}
{We also evaluate the accuracy of our FFNB on the challenging CIFAR100 dataset which includes 60k images belonging to 100 categories; 50k images are used for training and 10k for testing.  We use EfficientNet \cite{TL19} as our CNN backbone. In all the results in table~\ref{tab:cifar100_b50_s10a}, ``null-space + heteroscedasticity + multi-task initialization'' settings are used, and the band-size is set to $1$, number of layers in the feature map block to 2, size of minibatch to $32$, the learning rate fixed to 10e-3 and neither weight decay nor momentum are used. Performances are measured using the standard ``Average Incremental Acc'' which is proposed in iCaRL \cite{rebuffi2017icarl}, and defined as the average accuracy across all the visited tasks. Similarly to the standard evaluation protocol in iCaRL \cite{rebuffi2017icarl}, the first 50 classes ([1-50]) are used to pretrain the ``EfficientNet'' backbone, while the remaining 50 classes ([51-100]) are used for incremental task learning. Comparisons are shown in tables~\ref{tab:cifar100_b50_s10a} and~\ref{tab:cifar100_b50a} w.r.t. different tasks and related work. These results show that our proposed method provides a reasonable balance  between accuracy and the maximum number of training parameters w.r.t. these related methods.} 
\begin{table}
    \begin{center}
      \resizebox{0.99\textwidth}{!}{
        \begin{tabular}{l||*{11}{c|}c}

          \backslashbox{Test classes}{Tasks} & $\mathcal{T}_0$ & $\mathcal{T}_1$ & $\mathcal{T}_2$ & $\mathcal{T}_3$ & $\mathcal{T}_4$ & $\mathcal{T}_5$ & $\mathcal{T}_6$ & $\mathcal{T}_7$ & $\mathcal{T}_8$ & $\mathcal{T}_9$ & $\mathcal{T}_{10}$ & \textit{Average} \\
                                             & $[1-50]$ & $[51-55]$ & $[56-60]$ & $[61-65]$ & $[66-70]$ & $[71-75]$ & $[76-80]$ & $[81-85]$ & $[86-90]$ & $[91-95]$ & $[96-100]$ & \textit{Incremental Acc.}  \\
            \hline\hline
            Top-50 classes & 83.66 & 80.98 & 78.02 & 73.90 & 71.32 & 68.20 & 64.82 & 61.42 & 57.52 & 52.60 & 49.42 & 67.44 \\
            50 -- 55 & \multirow{10}{*}{--} & 79.20 & 71.60 & 70.40 & 65.80 & 58.00 & 56.60 & 43.80 & 43.60 & 34.00 & 26.00 & 54.90 \\
            55 -- 60 & & \multirow{9}{*}{--} & 89.40 & 84.00 & 74.20 & 71.40 & 68.60 & 63.80 & 61.80 & 60.20 & 57.00 & 70.04 \\
            60 -- 65 & & & \multirow{8}{*}{--} & 62.80 & 58.00 & 56.40 & 46.00 & 43.40 & 35.20 & 31.00 & 24.20 & 44.63 \\
            65 -- 70 & & & & \multirow{7}{*}{--} & 73.60 & 73.20 & 66.40 & 62.80 & 55.40 & 50.80 & 43.00 & 60.74 \\
            70 -- 75 & & & & & \multirow{6}{*}{--} & 80.60 & 76.40 & 68.00 & 59.40 & 47.80 & 41.60 & 62.30 \\
            75 -- 80 & & & & & & \multirow{5}{*}{--} & 80.80 & 80.20 & 71.20 & 67.60 & 62.40 & 72.44 \\
            80 -- 85 & & & & & & & \multirow{4}{*}{--} & 87.40 & 73.00 & 62.80 & 57.20 & 70.10 \\
            85 -- 90 & & & & & & & & \multirow{3}{*}{--} & 85.00 & 83.40 & 79.00 & 82.47 \\
            90 -- 95 & & & & & & & & & \multirow{2}{*}{--} & 82.60 & 74.60 & 78.60 \\
            95 -- 100 & & & & & & & & & & -- & 80.40 & 80.40 \\
            \hline
            \textit{Average Task Acc}. & 83.66  & 80.82 & 78.43 & 73.55 & 70.34 & 68.11 & 65.19 & 62.56 & 58.88 & 55.06 & 51.98 & 68.05 

        \end{tabular}}
    \end{center}
    \caption{\scriptsize {Results on \textit{CIFAR100-B50-S10}. Here $B50$ stands for the 50  pretraining classes and $S10$ for tasks $\T_1,\dots,\T_{10}$ which are learned incrementally (here $\T_1,\dots,\T_{10}$ correspond to classes [51-100] while $\T_0$ is the pretraining task involving classes [1-50]). In this table, the symbol ``--'' stands for ``accuracy not available'' as classes are incrementally visited so training+test data, belonging to the subsequent tasks, are obviously not available beforehand.}}\label{tab:cifar100_b50_s10a}
  \end{table}

\begin{table}
\centering
\parbox{0.4\textwidth}{
\begin{footnotesize}
 \scalebox{0.9}{ \begin{tabular}{l|*{5}{c|}c}

            Methods & \multicolumn{1}{c}{\#Params (M)} &  \multicolumn{1}{c}{Avg. Acc.} \\
            \hline\hline
            Upper Bound \cite{yan2021dynamically}  & \multicolumn{1}{c}{\textcolor{blue}{11.2}} & \multicolumn{1}{c}{\textcolor{blue}{79.91}} \\
            \hline
            iCaRL \cite{rebuffi2017icarl,yan2021dynamically} &  \multicolumn{1}{c}{\textcolor{blue}{11.2}} & \multicolumn{1}{c}{\textcolor{blue}{58.59}} \\
            UCIR  \cite{hou2019learning,yan2021dynamically} & \multicolumn{1}{c}{\textcolor{blue}{11.2}} & \multicolumn{1}{c}{\textcolor{blue}{59.92}} \\
            BiC  \cite{wu2019large,yan2021dynamically} & \multicolumn{1}{c}{\textcolor{blue}{11.2}} & \multicolumn{1}{c}{\textcolor{blue}{60.25}} \\
            WA   \cite{ZXG20,yan2021dynamically} & \multicolumn{1}{c}{\textcolor{blue}{11.2}} & \multicolumn{1}{c}{\textcolor{blue}{57.86}} \\
            PoDNet \cite{yan2021dynamically} & \multicolumn{1}{c}{\textcolor{blue}{11.2}} & \multicolumn{1}{c}{\textcolor{blue}{64.04} (63.19)} \\
            DDE (UCIR R20) \cite{HTM21} & \multicolumn{1}{c}{\textcolor{blue}{11.2}} & \multicolumn{1}{c}{62.36} \\
            DDE (PoDNet R20) \cite{HTM21} &\multicolumn{1}{c}{\textcolor{blue}{11.2}} & \multicolumn{1}{c}{64.12} \\
            DER(w/o P) \cite{yan2021dynamically} & \multicolumn{1}{c}{67.2} & \multicolumn{1}{c}{{72.81}} \\
            DER(P) \cite{yan2021dynamically} & \multicolumn{1}{c}{{8.79}} & \multicolumn{1}{c}{72.45} \\
            \hline
            Ours &  \multicolumn{1}{c}{{5.8}} & \multicolumn{1}{c}{{68.05}} \\

        \end{tabular}}

\end{footnotesize} \vspace{0.15cm}
\caption{\scriptsize {Results on \textit{CIFAR100-B50} (modified from Table 2 in DER \cite{yan2021dynamically} where numbers in \textcolor{blue}{blue} refer to the results tested by the re-implementation in DER \cite{yan2021dynamically} and numbers in parentheses refer to the results reported in the original papers).}}
\label{tab:cifar100_b50a}
}
\qquad
\begin{minipage}[c]{0.5\textwidth}
\centering
\centerline{\scalebox{0.43}{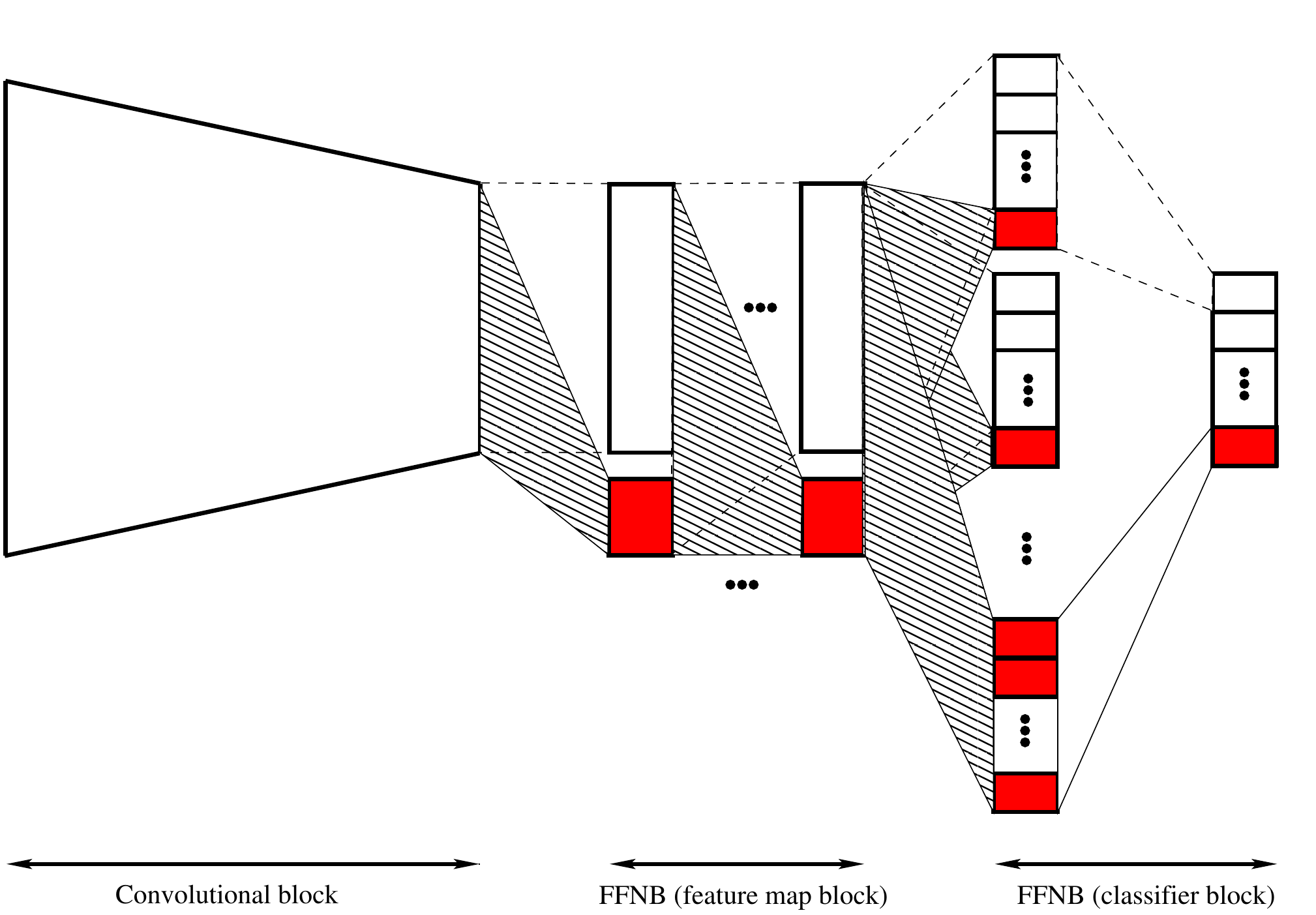}}
\caption{\scriptsize {This figure shows the whole architecture including a convolutional backbone and FFNB. For each new task  $\T_t$ (one or multiple new classes), bands of neurons (shown in red) are appended to the dynamic layers $\psi_1(.),\dots,\psi_{L-1}(.)$  of the feature map block, and only the underlying parameters (hatched) are trained in the null-space of the previous tasks. The classifier block includes two layers: in the first  layer, bands of neurons (in red) are appended to $\psi_{L}(.)$ in order to model all the ``one-vs-one'' classifiers involving the new classes and all the (previously and newly) visited classes so far. Finally, a band of neurons (again in red) is appended to the second classification layer $\psi_{L+1}(.)$ in order to aggregate the scores of the ``one-vs-one'' classifiers (see again \ref{FDAa}).}} 
\label{fig111111a}
\end{minipage}
\end{table}

\section{Conclusion}
We introduce  in this work a novel continual learning approach based on dynamic networks. The strength of the proposed method resides in its ability to learn discriminating representations and classifiers incrementally while providing stable behaviors on the previous tasks. The proposed method is based on FFNBs whose parameters are learned in the null-space of the previous tasks, leading to stable representations and classifications on these tasks. Aggregated classifiers are also learned incrementally using Fisher discriminant analysis which also exhibits optimal behavior especially when the feature maps are appropriately learned  and separable. Conducted experiments show the positive impact of each of the proposed components of our model.  As a future work, we are currently investigating multiple aspects including replay-based methods that allow fine-tuning the networks using not only {\it current} task data but also {\it previous} ones sampled from a generative network.
\newpage 

\section*{Appendix (Supplementary Material)}
This supplementary material includes the following items
\begin{itemize}
\item Incremental learning algorithms~1 and 2 (as discussed in section 3) of the paper.
\item Detailed proof of proposition~1.   
\item Experiments showing the impact of the number of FFNB-layers  on SBU and FPHA (tables~\ref{tableaa} and ~\ref{tablebb}).
\item Experiments showing the impact of the band-size allowed to each task in the feature maps of FFNB on SBU and FPHA (tables~\ref{tablehh} and~\ref{tableii}); see also section~\ref{baselines} in the paper.
\item Experiments showing the impact of pretraining the backbone network  +  FFNB fine-tuning on SBU and FPHA (tables~\ref{table6} and~\ref{table62}); see also section~\ref{ablation} in the paper. {Extra experiments are shown (in table~\ref{table6w}) w.r.t. increasing sizes of pretraining data.} 
\item Experiments showing a comparison between the aggregated ``one-vs-one'' classifiers w.r.t. the usual ``one-vs-all'' classifiers (tables~\ref{tablebbb} and~\ref{tablebbbb}); see also section~\ref{FDAa}.
\item Analysis of different factors intervening in the bound (in Eq.~\ref{eq3}): these factors correspond to dimension, weight decay regularization, and activation functions (tables~\ref{tableiii}, \ref{tableiii2}, \ref{tableiiii}, \ref{tableiiiii}, \ref{tableiiiibis} and \ref{tableiiiiibis}).
\item {Experiments showing the impact of the batch-normalization with and without our covariance normalization on SBU and FPHA (tables~\ref{table6aaaa} and \ref{table6bbbb}) + a detailed justification about the performances in the underlying caption.}
\item And, experiments on CIFAR100, including comparisons w.r.t. the closely related work (tables~\ref{tab:cifar100_b50_s2}, \ref{tab:cifar100_b50_s5}, \ref{tab:cifar100_b50_s10} and \ref{tab:cifar100_b50}). 

\end{itemize}   
\clearpage 
\newpage

\section*{Algorithms}

\begin{algorithm}[!ht]
\KwIn{Sequential tasks $\T_1,\dots\T_T$.}
\KwOut{Trained network parameters $\{\W_{\ell,t}\}_{t,\ell}$.}
\BlankLine
\For{$t:=1$ {\bf to} $T$}{
  \Repeat{convergence or max nbr of iterations reached}{
   {\bf Backward:} back-propagate $\frac{\partial E}{\partial \phii_{L+1}}$ and get $\frac{\partial E}{\partial \W_{\ell,t}}$;  \\  
  $\W_{\ell,t} \leftarrow \W_{\ell,t} - \nu \displaystyle \frac{\partial E}{\partial \W_{\ell,t}}$ \tcp*[r]{being $\nu$ the learning rate}
  Keep the parameters $\{\W_{\ell,r}\}_{r\neq t}$ of the other tasks unchanged;\\
  {\bf Forward:} update the outputs $\{\psi_\ell(\XX_t)\}_{\ell}$ on the current task $t$; 

  }
\BlankLine
}
 \caption{Incremental learning}\label{alg1}
\end{algorithm}

\begin{algorithm}[!ht]
\KwIn{Sequential tasks $\T_1,\dots\T_T$.}
\KwOut{Trained network parameters $\{\W_{\ell,t}\}_{t,\ell}$.}
\BlankLine
\For{$t:=1$ {\bf to} $T$}{

  Set $\{\P^\ell\}_\ell$ layerwise using PCA on the previous task outputs $\{\psi_\ell(\XX_\t)\}_{\ell}$. \\ 
  Set $\{\alpha_{\ell,t}\}_\ell$ using Eq.~(\ref{eq6})~or~(\ref{eq7}).   \\ 
  \Repeat{convergence or max nbr of iterations reached}{
  {\bf Backward:} back-propagate $\frac{\partial E}{\partial \phii_{L+1}}$ and get $\{\frac{\partial E}{\partial \W_{\ell,t}}\}_\ell$;  \\
    $\frac{\partial E}{\partial \alphaa_{\ell,t}} \leftarrow \frac{\partial E}{\partial \W_{\ell,t}} \ \frac{\partial \W_{\ell,t}}{\partial  \alphaa_{\ell,t}}$; \\
  $\alphaa_{\ell,t} \leftarrow \alphaa_{\ell,t} - \nu \displaystyle \frac{\partial E}{\partial \alphaa_{\ell,t}}, \forall \ell \in \{1,\dots,L\}$;  \\ 
  Update $\W_{\ell,t}$ using Eq.~(\ref{eq1}) and keep $\{\W_{\ell,r}\}_{r\neq t}$ of the other tasks unchanged; \\
  Update $\{\W_{L,(t,r)}\}_{r \in \t}$ using Eq.~(\ref{eq111}) and keep the others unchanged; \\
 {\bf Forward:} update the outputs $\{\psi_\ell(\XX_t)\}_{\ell}$ on the current task $t$; 
  }
\BlankLine
}
\caption{Updated incremental learning}\label{alg2}
\end{algorithm}
\clearpage 
\newpage
\section*{Analysis on SBU and FPHA}
In all the following tables, ``null-space + heteroscedasticity + multi-task initialization'' settings are used (following the ablation study in tables~\ref{tablecc} and~\ref{tabledd} in the paper). The number of FFNB feature layers and band-sizes are set to 3 (excepting particular settings in tables~\ref{tableaa},~\ref{tablebb} and \ref{tablehh}, \ref{tableii} respectively) and FFNB activations correspond to ReLU (excepting particular settings in  tables~\ref{tableiii} and~\ref{tableiii2}). Pretraining and fine-tuning is always used (excepting particular settings in  tables~\ref{table6} and~\ref{table62}). Our aggregated ``one-vs-one'' classifiers are also used in all these tables (excepting particular settings in  tables~\ref{tablebbb} and~\ref{tablebbbb}). \\

\noindent Note that accuracy when handling the first task in SBU is necessarily equal to 100\% as the first task includes only one class, while in FPHA the first task includes 5 classes, so the accuracy is lower (see again captions of tables~\ref{table1},~\ref{table12},~\ref{tablecc} and~\ref{tabledd} in the paper). 
\begin{table}[h]
  \begin{center}
\resizebox{0.9\textwidth}{!}{\begin{tabular}{c||cccccccc}
  \backslashbox{\# layers}{Tasks}    & $\T_1$ & $\T_2$ & $\T_3$ & $\T_4$ & $\T_5$ & $\T_6$ & $\T_7$ & $\T_8$  \\ 
  \hline
  \hline
  2 Layers  &    100.00&  100.00&  96.42&89.47&81.39&  79.16&   72.41& 70.76 \\
  3 Layers      & 100.00& 100.00& 100.00& 100.00& 97.67& 89.58& 89.65& \bf84.61   \\
 4 Layers      &  100.00 &  100.00&   96.42&   97.36&   93.02&   93.75&   75.86&   76.92 
 \end{tabular}}
\end{center}\caption{Impact of the number of layers in the FFNB-features on the performances  using SBU (in these experiments, $p=45$).} \label{tableaa}
\end{table}
\begin{table}[h]
  \begin{center}
\resizebox{0.95\textwidth}{!}{\begin{tabular}{c||ccccccccc}
  \backslashbox{\# layers}{Tasks}    & $\T_1$ & $\T_2$ & $\T_3$ & $\T_4$ & $\T_5$ & $\T_6$ & $\T_7$ & $\T_8$ & $\T_9$  \\ 
  \hline
  \hline
  3 Layers     & 68.25  & 58.73  & 64.76  & 63.70 &  62.84  & 61.24 &  64.96  & 66.01  & \bf67.47 \\   
 4 Layers      &  53.96  & 44.44 &  35.75  & 40.15 & 46.13 &  44.96 & 46.78 &  45.50 &  46.95  
 \end{tabular}}
\end{center}\caption{Impact of the number of layers in the FFNB-features on the performances using FPHA  (in these experiments, $p=75$).
} \label{tablebb}
\end{table}
 \begin{table}[h]
   \begin{center}
 \resizebox{0.95\textwidth}{!}{\begin{tabular}{c||cccccccc}
   \backslashbox{Band size}{Tasks}    & $\T_1$ & $\T_2$ & $\T_3$ & $\T_4$ & $\T_5$ & $\T_6$ & $\T_7$ & $\T_8$  \\ 
   \hline
   \hline 
   1     &  100.00 &100.00 &100.00& 92.10& 86.04& 87.50& 82.75& 64.61    \\
    3     &  100.00& 100.00& 100.00& 100.00& 97.67& 89.58& 89.65& \bf84.61\\
    5     &  100.00& 100.00& 100.00& 100.00& \bf100.00& \bf95.83& \bf91.37& 81.53     \\
    7     &   100.00& 100.00& 100.00& 94.73& 90.69& 89.58& 77.58& 67.69   \\
    9     & 100.00 &95.00& 92.85& 94.73& 93.02& 89.58& 82.75& 73.84      
 \end{tabular}}
 \end{center}\caption{Impact of band-size on the performances using SBU (again $p=45$).}  \label{tablehh}
\end{table}

\begin{table}[h]
  \begin{center}
\resizebox{0.95\textwidth}{!}{\begin{tabular}{c||ccccccccc}
  \backslashbox{Band size}{Tasks}   & $\T_1$ & $\T_2$ & $\T_3$ & $\T_4$ & $\T_5$ & $\T_6$ & $\T_7$ & $\T_8$ & $\T_9$  \\  
  \hline
  \hline 
  1     & 68.25 &  55.55  & 60.62 &  62.16 &  59.75 & 60.46 &  61.86 &  62.10  & 62.43 \\ 
  3     & 68.25  & 58.73  & 64.76  & 63.70 &  62.84  & 61.24 &  64.96  & 66.01  & \bf67.47 \\ 
  5     &  63.49 &  53.17 &  60.10 &  57.52 &  57.58 &  56.58 &  60.31 &  61.91 &  61.91 \\ 
                                7    &  58.73&   50.00&   58.54 & 61.00&   59.13&   52.97&   55.87&   59.96&  60.17  \\ 
                                9 &  61.90 &  51.78 &  51.38  & 53.87 &  51.93 &  58.13 &  60.22 &  62.32 &  62.69 
\end{tabular}}
\end{center}\caption{Impact of band-size on the performances using FPHA ($p=75$).}  \label{tableii}
\end{table}

\begin{table}[h]
  \begin{center}
    \resizebox{1\textwidth}{!}{
    \begin{tabular}{cc|cccccccc}
   Pretraining & Fine-tuning  &   $\T_1$ & $\T_2$ & $\T_3$ & $\T_4$ & $\T_5$ & $\T_6$ & $\T_7$ & $\T_8$ \\
      \hline
      \hline 
   \xmark   & \xmark &  100.00&  100.00&  97.36&  95.34&  93.75&  87.93&   83.07&  83.07     \\
   \xmark   & \cmark &   100.00&  100.00&   97.36&   93.02&   87.50&   87.93&   84.61&   \bf84.61  \\ 
      \cmark  & \xmark &  100.00&  100.00&  100.00&   97.36&  90.69&  87.50&   87.93&  83.07   \\
      \cmark & \cmark &   100.00&   100.00&    100.00&   \bf100.00 &   \bf97.67  &   \bf89.58  &   \bf89.65  &   \bf84.61  
\end{tabular}}
\end{center}\caption{Impact of pretraining and fine-tuning on the performances using SBU (here $p=45$).}   \label{table6}
\end{table}

\begin{table}[h]
  \begin{center}
    \resizebox{1\textwidth}{!}{
    \begin{tabular}{cc|ccccccccc}
   Pretraining & Fine-tuning  &   $\T_1$ & $\T_2$ & $\T_3$ & $\T_4$ & $\T_5$ & $\T_6$ & $\T_7$ & $\T_8$& $\T_9$ \\
      \hline
      \hline 
   \xmark   & \xmark &   63.15 &  59.52 &  59.58 &  58.30 &  58.82 &  57.10  & 60.75 & 62.89 &  63.13 \\ 
   \xmark   & \cmark &   64.47 & 60.86 &  62.43 &  56.08 &  58.20 &  55.13 &  58.74 &  58.96 &  60.34 \\ 
      \cmark  & \xmark &  68.25 &  59.52 &  61.65  & 61.77 &  59.13 &  58.39 &  60.31 &  61.52 &  63.30 \\
      \cmark & \cmark &  68.25  & 58.73  & 64.76  & 63.70 &  62.84  & 61.24 &  64.96  & 66.01  & \bf67.47 
\end{tabular}}
\end{center}\caption{Impact of pretraining and fine-tuning on the performances using FPHA (here $p=75$).}   \label{table62}
\end{table}

\begin{table}[h]
  \begin{center}
\resizebox{0.99\textwidth}{!}{\begin{tabular}{c||cccccccc}
  \backslashbox{FFNB classifiers}{Tasks}    & $\T_1$ & $\T_2$ & $\T_3$ & $\T_4$ & $\T_5$ & $\T_6$ & $\T_7$ & $\T_8$ \\ 
  \hline
  \hline
``One-vs-all''  classifiers    & 100.00 & 100.00 & 82.14 & 81.57 & 65.11 & 66.66 & 60.34 & 55.38 \\ 
Our aggregated ``one-vs-one'' classifiers &  100.00&  100.00&   97.36&   93.02&   87.50&   87.93&   84.61&   \bf84.61 \\ 
 \end{tabular}}
\end{center}\caption{Impact of FFNB-classifiers on the performances using SBU  (in these experiments, $p=45$).} \label{tablebbb}
\end{table}

\begin{table}[h]
  \begin{center}
\resizebox{0.99\textwidth}{!}{\begin{tabular}{c||ccccccccc}
  \backslashbox{FFNB classifiers}{Tasks}    & $\T_1$ & $\T_2$ & $\T_3$ & $\T_4$ & $\T_5$ & $\T_6$ & $\T_7$ & $\T_8$ & $\T_9$  \\ 
  \hline
  \hline
``One-vs-all'' classifiers   & 38.09 &  29.36&   47.66 &   43.62 &   36.53 &  36.17 &  35.25 &  35.74 &  35.65 \\ 
 Our aggregated ``one-vs-one'' classifiers &  68.25  & 58.73  & 64.76  & 63.70 &  62.84  & 61.24 &  64.96  & 66.01  & \bf67.47 \\ 
 \end{tabular}}
\end{center}\caption{Impact of FFNB-classifiers on the performances using FPHA  (in these experiments, $p=75$).} \label{tablebbbb}
\end{table}

\begin{table}[h]
  \begin{center}
\resizebox{0.8\textwidth}{!}{\begin{tabular}{c||c|c}
  \backslashbox{FFNB-Activations}{Performances}   & CF Bound in Eq~(\ref{eq3}) & Accuracy (@final task $\T_8$)  \\  
  \hline
  \hline
Tanh           &  0.1581 &   72.30  \\
Sigmoid       &    5.75 $\times 10^{-5}$ &   76.92  \\
ReLU           &  15.98 $\times 10^{-7}$   & \bf84.61 \\
                              \end{tabular}}
\end{center}\caption{Impact of activations  on the catastrophic forgetting (CF) bound and performances using SBU ($p=45$).}  \label{tableiii}
\end{table}
\begin{table}[h]
  \begin{center}
\resizebox{0.8\textwidth}{!}{\begin{tabular}{c||c|c}
  \backslashbox{FFNB-Activations}{Performances}   &  CF Bound in Eq~(\ref{eq3})  &  Accuracy (@final task $\T_9$)  \\  
  \hline
  \hline
Tanh           &  19.23 &   60.86  \\
Sigmoid       &   18.52  &  60.86  \\
ReLU           &  0.288   & \bf67.47 \\
                              \end{tabular}}
\end{center}\caption{Impact of activations  on the CF bound and performances using FPHA ($p=75$).}  \label{tableiii2}
\end{table}

\begin{table}[h]
  \begin{center}
\resizebox{0.8\textwidth}{!}{\begin{tabular}{c||c|c}
  \backslashbox{Dimensions}{Performances}   & CF Bound in Eq~(\ref{eq3}) ($\times 10^{-7}$)     &  Accuracy (@final task $\T_8$)  \\  
  \hline
  \hline 
$p=15$      &  25618.3 & 81.53 \\
$p=25$      &  437.39    & 83.07 \\
$p=35$      &  64.12    & 81.53 \\
$p=45$      &  15.98    & \bf84.61 \\
$p=50$      &  7.80    & 73.84 \\
$p=55$      &  2.54    & 69.23 \\                              
                              \end{tabular}}
\end{center}\caption{Impact of dimensions ($p$)  on the CF bound and performances using SBU (with ReLU). From these results, small $p$ leads to CF while large $p$ to low dimensional and noisy null-space (and hence low generalization), so the best performances are obtained when sufficiently (but not very) large $p$-values are selected.}  \label{tableiiii}
\end{table}

\begin{table}[h]
  \begin{center}
\resizebox{0.8\textwidth}{!}{\begin{tabular}{c||c|c}
  \backslashbox{Dimensions}{Performances}   & CF Bound in Eq~(\ref{eq3})   &  Accuracy (@final task $\T_9$)   \\  
  \hline
  \hline 
$p=15$      & 425.716411    &  63.82  \\
$p=30$      &   11.005551    &  61.56  \\
$p=45$      &   2.656760   & 63.30 \\
$p=60$      &   0.836268    &  64.69  \\
$p=75$      &  0.288356   &  \bf67.47 \\
$p=90$      &    0.115226  & 60.69 \\
$p=105$     &   0.043147   & 60.00 \\
$p=120$     &    0.015686  & 57.21 \\
\end{tabular}}
\end{center}\caption{Impact of dimensions ($p$) on the CF bound and performances using FPHA (with ReLU). Again, from these results, small $p$ leads to CF while large $p$ to low dimensional and noisy null-space (and hence low generalization), so the best performances are obtained when sufficiently (but not very) large $p$-values are selected.}  \label{tableiiiii}
\end{table}

\begin{table}[h]
  \begin{center}
\resizebox{0.85\textwidth}{!}{\begin{tabular}{c||c|c}
  \backslashbox{Weight decay coefficient}{Performances}   & CF Bound in Eq~(\ref{eq3})  ($\times 10^{-7}$)   & Accuracy (@final task $\T_8$)  \\  
  \hline
  \hline
$10^{-8}$      &  15.98   & \bf84.61 \\                   
$10^{-7}$      &  15.86   & 83.07 \\
$10^{-6}$      &  14.89   & 81.53 \\
$10^{-5}$      &  7.59    & 78.46 \\                                
$10^{-4}$      &  6.59    & 78.46 \\
$10^{-3}$      &  4.51    & 81.53 \\
$10^{-2}$      &  2.10    &  76.92 \\                              
                              \end{tabular}}
\end{center}\caption{Impact of weight decay regularization on the CF bound and performances using SBU (with ReLU and $p=45$).}  \label{tableiiiibis}
\end{table}

\begin{table}[h]
  \begin{center}
\resizebox{0.85\textwidth}{!}{\begin{tabular}{c||c|c}
  \backslashbox{Weight decay coefficient}{Performances}   & CF Bound in Eq~(\ref{eq3})  &  Accuracy (@final task $\T_9$)  \\  
  \hline
  \hline
$10^{-8}$      &   0.288356  & \bf67.47 \\                   
$10^{-7}$      &  0.281536   & 64.00 \\
$10^{-6}$      &  0.238607   & 64.17 \\
$10^{-5}$      &  0.170736   & 55.13 \\                                
$10^{-4}$      &  0.134717   & 55.82 \\
$10^{-3}$      &  0.116821   & 47.65 \\
$10^{-2}$      &  0.139448    &  47.47 \\                              
                              \end{tabular}}
\end{center}\caption{Impact of weight decay regularization on the CF bound and performances using FPHA (with ReLU and $p=75$).}  \label{tableiiiiibis}
\end{table}

\begin{table}[h]
  \begin{center}
    \resizebox{1\textwidth}{!}{
    \begin{tabular}{cc|cccccccc}
   Batch-norm & Heteroscedasticity  &   $\T_1$ & $\T_2$ & $\T_3$ & $\T_4$ & $\T_5$ & $\T_6$ & $\T_7$ & $\T_8$ \\
      \hline
      \hline 
   \xmark   & \xmark &  100.00 & 100.00 & 100.00 &  97.36 & 90.69&   87.50&   82.75&  75.38 \\
      \xmark   & \cmark &  100.00 &  100.00 & 100.00 &  100.00 &  \bf97.67 &  89.58 &  89.65 &  \bf84.61 \\                           
      \cmark  & \xmark &  100.00 &  85.00&   67.85 &  55.26 &  41.86 & 33.33 &  32.75 &  29.23  \\
      \cmark & \cmark & 100.00 & 100.00&  96.42&  100.00&   95.34 &  \bf95.83 &  \bf93.10 &  83.07   
\end{tabular}}
\end{center}\caption{{Impact of batch-norm (BN), with and w/o our class-wise covariance normalization, on SBU (here $p=45$). The reason why BN is degrading performances is not intrinsically related to the BN itself (which is known to be effective in the general multi-task setting), but due to the incremental setting (i.e., due to the interference introduced by the BN on the previous tasks; put differently, the feature maps of the FFNB network on the previous tasks are no longer guaranteed to belong to the residual space when BN is applied).}}   \label{table6aaaa}
\end{table}

\begin{table}[h]
  \begin{center}
    \resizebox{1\textwidth}{!}{
\begin{tabular}{cc|ccccccccc}
 Batch-norm & Heteroscedasticity  &   $\T_1$ & $\T_2$ & $\T_3$ & $\T_4$ & $\T_5$ & $\T_6$ & $\T_7$ & $\T_8$ & $\T_9$ \\
      \hline
      \hline 
   \xmark   & \xmark   &  76.19 & 67.46  &  64.76  & 62.93 &  62.84 & 61.24  &  59.42  & 58.00  &  54.95  \\
   \xmark   & \cmark   &  68.25 & 58.73  &  64.76  & \bf63.70 &  62.84 & 61.24  &  \bf64.96  & \bf66.01  &  \bf67.47  \\ 
      \cmark  & \xmark &  52.38 & 45.23  &  34.71  & 32.81 &  36.22 & 39.53  &  39.24  & 45.31  &  49.21 \\
      \cmark & \cmark  &  74.60 & 57.93  &  33.16  & 44.01 &  48.60 & 53.74  &  58.09  & 58.98  &  57.39  
\end{tabular}}
\end{center}\caption{{Impact of batch normalization (with and wo covariance normalization) on FPHA (here $p=75$). We observe a similar behavior as SBU (see caption of the previous table).}}   \label{table6bbbb}
\end{table}

\begin{table}[h]
  \begin{center}
    \resizebox{1\textwidth}{!}{
    \begin{tabular}{c|ccccccccc}
 Configuration    &   $\T_1$ & $\T_2$ & $\T_3$ & $\T_4$ & $\T_5$ & $\T_6$ & $\T_7$ & $\T_8$ & $\T_9$\\
      \hline
      \hline 
                 no-pretraining  &  63.15  & 59.52  &  59.58   & 58.30  & 58.82&   57.10 &  60.75  & 62.89 &  63.13    \\
 pretraining (25\% of pretraining data)&  68.25 &  49.20 &  55.95 &  58.30 &  63.46 &  61.49 & 62.97 &  62.69 &  63.47 \\              
                 pretraining (50\% of pretraining data) &  57.14 &  56.34  &  63.21 &   61.77 &  60.68 &  62.27 &  63.85 &  63.28  & 63.82  \\

   pretraining (100\% of pretraining data)  &  68.25 &  58.73  &  64.76  &  63.70 &  62.84 &  61.24 &  64.96  & 66.01  & 67.47 
\end{tabular}}
\end{center}\caption{Impact of backbone pretraining on the performances using FPHA (here $p=75$). Results shown in this table provide an idea about the behavior of our FFNB w.r.t. increasing pretraining sets and also w.r.t. no-pretraining of the backbone. In spite of no-pretraining, FFNB (also endowed with feature map layers) is able to adapt the features to the new incremental tasks prior to achieve classification. This is possible thanks to the new dynamic parameters of the current task which are trained in the null-space of the previous tasks, and this  mitigates CF.}\label{table6w}
\end{table}

\clearpage 
\newpage

\section*{Evaluation on CIFAR100 and SOTA Comparison}
In all the following tables, ``null-space + heteroscedasticity + multi-task initialization'' settings are used. On CIFAR100, the band-size=1, size of minibatch=32, optimizer=SGD, learning rate fixed to 10e-3 and neither weight decay nor momentum are used. 

\begin{table}[h]
    \begin{center}
         \resizebox{0.79\textwidth}{!}{
        \begin{tabular}{l||*{3}{c|}c}
            \hline
            \backslashbox{Test classes}{Tasks} & $\mathcal{T}_0$ & $\mathcal{T}_1$ & $\mathcal{T}_2$ & \textit{Average Incremental Acc}. \\
            \hline\hline
            Top-50 classes & 83.66 & 76.18 & 68.40 & 76.08 \\
            50 -- 75 & \multirow{2}{*}{--} & 60.96 & 51.52 & 56.24 \\
            75 -- 100 & & -- & 60.68 & 60.68 \\
            \hline
            \textit{Average Task Acc}. & 83.66 & 71.11 & 62.25 & 72.34 \\
            \hline
        \end{tabular}
        }
    \end{center}
    \caption{Results on \textit{CIFAR100-B50-S2}. As suggested by the standard evaluation protocol, the first 50 classes ([1-50]) are used to pretrain the ``EfficientNet'' backbone, while the remaining 50 classes ([51-100]) are used for incremental task learning. Here $B50$ stands for these 50  pretraining classes and $S2$ for tasks $\T_1$ and $\T_2$ which are learned incrementally (here $\T_1$ corresponds to classes [51-75] and $\T_2$  to [76-100] while $\T_0$ is the pretraining task involving classes [1-50]). \textit{Average Incremental Acc} is proposed in iCaRL \textrm{[RKSL17]} which is averaged across tasks. The symbol ``--'' stands for ``accuracy not available'' as classes are incrementally visited so training+test data, belonging to the subsequent tasks, are obviously not available beforehand.}
    \label{tab:cifar100_b50_s2}
\end{table}

\begin{table}[h]
    \begin{center}
       \resizebox{0.99\textwidth}{!}{
        \begin{tabular}{l||*{6}{c|}c}
            \hline
            \backslashbox{Test classes}{Tasks} & $\mathcal{T}_0$ & $\mathcal{T}_1$ & $\mathcal{T}_2$ & $\mathcal{T}_3$ & $\mathcal{T}_4$ & $\mathcal{T}_5$ & \textit{Average Incremental Acc}. \\
            \hline\hline
            Top-50 classes & 83.66 & 79.26 & 74.74 & 70.76 & 64.76 & 58.18 & 71.89 \\
            50 -- 60 & \multirow{5}{*}{--} & 76.80 & 71.30 & 67.00 & 58.20 & 50.30 & 64.72 \\
            60 -- 70 & & \multirow{4}{*}{--} & 63.10 & 57.10 & 50.00 & 41.50 & 52.93 \\
            70 -- 80 & & & \multirow{3}{*}{--} & 69.10 & 61.10 & 52.60 & 60.73 \\
            80 -- 90 & & & & \multirow{2}{*}{--} & 75.50 & 69.00 & 72.25 \\
            90 -- 100 & & & & & -- & 70.90 & 70.90 \\
            \hline
            \textit{Average Task Acc}. & 83.66 & 78.55 & 72.59 & 68.38 & 63.18 & 57.52 & 70.70 \\
            \hline
        \end{tabular}
        }
    \end{center}
    \caption{Results on \textit{CIFAR100-B50-S5}. The caption of this table is similar to table \ref{tab:cifar100_b50_s2} excepting the number of tasks which is now equal to 5.}
    \label{tab:cifar100_b50_s5}
\end{table}

\begin{table}[h]
    \begin{center}
      \resizebox{0.99\textwidth}{!}{
        \begin{tabular}{l||*{11}{c|}c}
            \hline
            \backslashbox{Test classes}{Tasks} & $\mathcal{T}_0$ & $\mathcal{T}_1$ & $\mathcal{T}_2$ & $\mathcal{T}_3$ & $\mathcal{T}_4$ & $\mathcal{T}_5$ & $\mathcal{T}_6$ & $\mathcal{T}_7$ & $\mathcal{T}_8$ & $\mathcal{T}_9$ & $\mathcal{T}_{10}$ & \textit{Average Incremental Acc}. \\
            \hline\hline
            Top-50 classes & 83.66 & 80.98 & 78.02 & 73.90 & 71.32 & 68.20 & 64.82 & 61.42 & 57.52 & 52.60 & 49.42 & 67.44 \\
            50 -- 55 & \multirow{10}{*}{--} & 79.20 & 71.60 & 70.40 & 65.80 & 58.00 & 56.60 & 43.80 & 43.60 & 34.00 & 26.00 & 54.90 \\
            55 -- 60 & & \multirow{9}{*}{--} & 89.40 & 84.00 & 74.20 & 71.40 & 68.60 & 63.80 & 61.80 & 60.20 & 57.00 & 70.04 \\
            60 -- 65 & & & \multirow{8}{*}{--} & 62.80 & 58.00 & 56.40 & 46.00 & 43.40 & 35.20 & 31.00 & 24.20 & 44.63 \\
            65 -- 70 & & & & \multirow{7}{*}{--} & 73.60 & 73.20 & 66.40 & 62.80 & 55.40 & 50.80 & 43.00 & 60.74 \\
            70 -- 75 & & & & & \multirow{6}{*}{--} & 80.60 & 76.40 & 68.00 & 59.40 & 47.80 & 41.60 & 62.30 \\
            75 -- 80 & & & & & & \multirow{5}{*}{--} & 80.80 & 80.20 & 71.20 & 67.60 & 62.40 & 72.44 \\
            80 -- 85 & & & & & & & \multirow{4}{*}{--} & 87.40 & 73.00 & 62.80 & 57.20 & 70.10 \\
            85 -- 90 & & & & & & & & \multirow{3}{*}{--} & 85.00 & 83.40 & 79.00 & 82.47 \\
            90 -- 95 & & & & & & & & & \multirow{2}{*}{--} & 82.60 & 74.60 & 78.60 \\
            95 -- 100 & & & & & & & & & & -- & 80.40 & 80.40 \\
            \hline
            \textit{Average Task Acc}. & 83.66  & 80.82 & 78.43 & 73.55 & 70.34 & 68.11 & 65.19 & 62.56 & 58.88 & 55.06 & 51.98 & 68.05 \\
            \hline
        \end{tabular}}
    \end{center}
    \caption{Results on \textit{CIFAR100-B50-S10}. The caption of this table is similar to table \ref{tab:cifar100_b50_s2} excepting the number of tasks which is now equal to 10.}
    \label{tab:cifar100_b50_s10}
  \end{table}

\begin{table}[h]
    \begin{center}
           \resizebox{0.99\textwidth}{!}{
        \begin{tabular}{l|*{5}{c|}c}
            \hline
            \multirow{2}{*}{Methods} & \multicolumn{2}{c}{2 tasks ($S2$)} & \multicolumn{2}{|c}{5 tasks ($S5$)} & \multicolumn{2}{|c}{10 tasks ($S10$)} \\
            & \multicolumn{1}{c}{\#Params (M)} & \multicolumn{1}{c|}{Avg. Acc.} & \multicolumn{1}{c}{\#Params (M)} & \multicolumn{1}{c|}{Avg. Acc.} &  \multicolumn{1}{c}{\#Params (M)} & \multicolumn{1}{c}{Avg. Acc.} \\
            \hline\hline
            Upper Bound [YXH21]  & \multicolumn{1}{c}{\textcolor{blue}{11.2}} & \multicolumn{1}{c|}{67.38 / \textcolor{blue}{72.22}} & \multicolumn{1}{c}{\textcolor{blue}{11.2}} & \multicolumn{1}{c|}{\textcolor{blue}{79.89}} & \multicolumn{1}{c}{\textcolor{blue}{11.2}} & \multicolumn{1}{c}{\textcolor{blue}{79.91}} \\
            \hline
            iCaRL[RKSL17,YXH21] & \multicolumn{1}{c}{\textcolor{blue}{11.2}} & \multicolumn{1}{c|}{\textcolor{blue}{71.33}} & \multicolumn{1}{c}{\textcolor{blue}{11.2}} & \multicolumn{1}{c|}{\textcolor{blue}{65.06}} & \multicolumn{1}{c}{\textcolor{blue}{11.2}} & \multicolumn{1}{c}{\textcolor{blue}{58.59}} \\
            UCIR [HPL+19,YXH21] & \multicolumn{1}{c}{\textcolor{blue}{11.2}} & \multicolumn{1}{c|}{\textcolor{blue}{67.21}} & \multicolumn{1}{c}{\textcolor{blue}{11.2}} & \multicolumn{1}{c|}{\textcolor{blue}{64.28}} & \multicolumn{1}{c}{\textcolor{blue}{11.2}} & \multicolumn{1}{c}{\textcolor{blue}{59.92}} \\
            BiC [WCW+19,YXH21] & \multicolumn{1}{c}{\textcolor{blue}{11.2}} & \multicolumn{1}{c|}{\textcolor{blue}{72.47}} & \multicolumn{1}{c}{\textcolor{blue}{11.2}} & \multicolumn{1}{c|}{\textcolor{blue}{66.62}} & \multicolumn{1}{c}{\textcolor{blue}{11.2}} & \multicolumn{1}{c}{\textcolor{blue}{60.25}} \\
            WA [ZXG+20,YXH21] & \multicolumn{1}{c}{\textcolor{blue}{11.2}} & \multicolumn{1}{c|}{\textcolor{blue}{71.43}} & \multicolumn{1}{c}{\textcolor{blue}{11.2}} & \multicolumn{1}{c|}{\textcolor{blue}{64.01}} & \multicolumn{1}{c}{\textcolor{blue}{11.2}} & \multicolumn{1}{c}{\textcolor{blue}{57.86}} \\
            PoDNet [YXH21] & \multicolumn{1}{c}{\textcolor{blue}{11.2}} & \multicolumn{1}{c|}{\textcolor{blue}{71.30}} & \multicolumn{1}{c}{\textcolor{blue}{11.2}} & \multicolumn{1}{c|}{\textcolor{blue}{67.25} (64.83)} & \multicolumn{1}{c}{\textcolor{blue}{11.2}} & \multicolumn{1}{c}{\textcolor{blue}{64.04} (63.19)} \\
            DDE (UCIR R20) [HTM+21] & \multicolumn{1}{c}{-} & \multicolumn{1}{c|}{\textbf{-}} & \multicolumn{1}{c}{\textcolor{blue}{11.2}} & \multicolumn{1}{c|}{65.27} & \multicolumn{1}{c}{\textcolor{blue}{11.2}} & \multicolumn{1}{c}{62.36} \\
            DDE (PoDNet R20) [HTM+21] & \multicolumn{1}{c}{-} & \multicolumn{1}{c|}{\textbf{-}} & \multicolumn{1}{c}{\textcolor{blue}{11.2}} & \multicolumn{1}{c|}{65.42} & \multicolumn{1}{c}{\textcolor{blue}{11.2}} & \multicolumn{1}{c}{64.12} \\
            DER(w/o P) [YXH21] & \multicolumn{1}{c}{22.4} & \multicolumn{1}{c|}{{74.61}} & \multicolumn{1}{c}{39.2} & \multicolumn{1}{c|}{{73.21}} & \multicolumn{1}{c}{67.2} & \multicolumn{1}{c}{{72.81}} \\
            DER(P) [YXH21] & \multicolumn{1}{c}{{3.90}} & \multicolumn{1}{c|}{74.57} & \multicolumn{1}{c}{{6.13}} & \multicolumn{1}{c|}{72.60} & \multicolumn{1}{c}{{8.79}} & \multicolumn{1}{c}{72.45} \\
            \hline
            Ours & \multicolumn{1}{c}{{5.8}} & \multicolumn{1}{c|}{{72.34}} & \multicolumn{1}{c}{{5.8}} & \multicolumn{1}{c|}{{70.70}} & \multicolumn{1}{c}{{5.8}} & \multicolumn{1}{c}{{68.05}} \\
            \hline 
        \end{tabular}
        }
    \end{center}
    \caption{Results on \textit{CIFAR100-B50} (modified from Table 2 in DER [YXH21] where numbers in \textcolor{blue}{blue} refer to the results tested by the re-implementation in DER [YXH21] and numbers in parentheses refer to the results reported in the original papers).}
    \label{tab:cifar100_b50}
  \end{table}

We use EfficientNet \textrm{[TL19]} as our backbone which is state-of-the-art feature extractor architecture. Comparison shown in table~\ref{tab:cifar100_b50} are w.r.t. the following related work 
\begin{itemize}  \small 

\item  \textrm{[HPL+19]} Saihui Hou, Xinyu Pan, Chen Change Loy, Zilei Wang, and Dahua Lin. Learning a unified classifier incrementally via rebalancing. In Proceedings of the IEEE/CVF Conference on Computer Vision and Pattern Recognition, pages 831–839, 2019. 
\item  \textrm{[HTM+21]} Xinting Hu, Kaihua Tang, Chunyan Miao, Xian-Sheng Hua, and Hanwang Zhang. Distilling causal effect of data in class-incremental learning. arXiv preprint arXiv:2103.01737, 2021. 
\item  \textrm{[RKSL17]} Sylvestre-Alvise Rebuffi, Alexander Kolesnikov, Georg Sperl, and Christoph H Lampert. icarl: Incremental classifier and representation learning. In Proceedings of the IEEE Conference on Computer Vision and Pattern Recognition, pages 2001–2010, 2017. 
\item  \textrm{[TL19]} Mingxing Tan and Quoc Le. Efficientnet: Rethinking model scaling for convolutional neural networks. In International Conference on Machine Learning, pages 6105–6114. PMLR, 2019. 
\item  \textrm{[WCW+19]} Yue Wu, Yinpeng Chen, Lijuan Wang, Yuancheng Ye, Zicheng Liu, Yandong Guo, and Yun Fu. Large scale incremental learning. In Proceedings of the IEEE/CVF Conference on Computer Vision and Pattern Recognition, pages 374–382, 2019. 
\item  \textrm{[YXH21]} Shipeng Yan, Jiangwei Xie, and Xuming He. Der: Dynamically expandable representation for class incremental learning. arXiv preprint arXiv:2103.16788, 2021.
  \item \textrm{[ZXG+20]} Bowen Zhao, Xi Xiao, Guojun Gan, Bin Zhang, and Shu-Tao Xia. Maintaining discrimination and fairness in class incremental learning. In Proceedings of the IEEE/CVF Conference on Computer Vision and Pattern Recognition, pages 13208–13217, 2020.
 \end{itemize} 

\clearpage 
\newpage
\section*{Proposition 1}
\begin{Proposition}
  Let $g: \mathbb{R} \rightarrow \mathbb{R}$ be a $L$-Lipschitz continuous activation (with $L\leq 1$). Any $\eta$-step descent (update) of  $\W_{\ell,t}$ in $\N_S(\phii_\ell(\XX_\t))$ using (\ref{eq1}) satisfies $\forall r \in \t$
  \begin{equation}\label{eq333} \small
    \begin{array}{lll}
      \big\|\phii_\ell^{\eta}({\XX}_r)- \phii_\ell^0({\XX_r})  \big\|_F^2  \leq B \\
     \\

  \textrm{with}  \ \ \  \ \ \  \ \ \       B = \displaystyle \sum_{\tau=1}^{\eta}  \sum_{k=0}^{\ell-1}   \big(\big\| \alpha_{\ell-k,t}^\tau\big\|_F^2 . \big\| \beta_{\ell-k-1,r}^{\tau}   \big\|_F^2 +  \big\| {\alpha}_{\ell-k,t}^{\tau-1}\big\|_F^2  . \big\| {\beta}_{\ell-k-1,r}^{\tau-1} \big\|_F^2 \big). \prod_{k'=0}^{k-1}  \big\|\W_{\ell-k',\r}^{\tau} \big\|_F^2,
\end{array}
       \end{equation} 
being  $\phii_\ell^{0}({\XX}_r)$  (resp. $\phii_\ell^{\eta-1}({\XX}_r)$) the map before the start (resp. the end) of the iterative update (descent on current task $\T_t$), $\beta_{\ell,r}^\tau$ the projection of $\phii_\ell^{\tau}({\XX}_r)$ onto $\N_S(\phii_\ell(\XX_\t))$ at any iteration $\tau$, $\{\W_{\ell,r}^\tau\}_\ell$ the network parameters at $\tau$, and $\|.\|_F$ the  Frobenius norm.
\end{Proposition} 

\subsection*{Proof of Proposition~1}                                                                                                                    {\footnotesize                                                                
At any iteration $\tau$ of the descent, one may write $\forall r \in \t$  
  \begin{eqnarray*} \small 
    \big\|\phii_\ell^\tau({\XX}_r) - \phii_\ell^{\tau-1}({\XX}_r)  \big\|_F^2 &=&     \big\|g\big(\W_{\ell}^\tau \ \phii_{\ell-1}^\tau({\XX}_r)\big)  - g\big({\W}_{\ell}^{\tau-1} \  \phii_{\ell-1}^{\tau-1}({\XX}_r)\big) \big\|_F^2  \\
      &  & \\
                                                                 & \leq &     \big\|\W_{\ell}^\tau \  \phii_{\ell-1}^\tau({\XX}_r)  - {\W}_{\ell}^{\tau-1} \  \phii_{\ell-1}^{\tau-1}({\XX}_r)  \big\|_F^2 \ \  \ \ \ \      ( g \  \textrm{$L$-Lipschitzian with} \ L\leq 1) \\
      &   & \\
                                                   &=&  \big\|\W_{\ell,t}^\tau \  \phii_{\ell-1}^\tau({\XX}_r)  - {\W}_{\ell,t}^{\tau-1} \   \phii_{\ell-1}^{\tau-1}({\XX}_r)  \big\|_F^2  \\
                                                                 & & \ \ \ \ \ \ \  +  \ \ \     \big\|\W_{\ell,r}^\tau \   \phii_{\ell-1}^\tau({\XX}_r)  - {\W}_{\ell,\r}^{\tau} \  \phii_{\ell-1}^{\tau-1}({\XX}_r)  \big\|_F^2  \ \  \  \ \ \   ({\W}_{\ell,\r}^{\tau-1}=\W_{\ell,\r}^{\tau}, \ \forall \tau)\\
     &  & \\
                                                   &\leq &  \big\|\W_{\ell,t}^{\tau} \   \phii_{\ell-1}^\tau({\XX}_r) - {\W}_{\ell,t}^{\tau-1} \   \phii_{\ell-1}^{\tau-1}({\XX}_r) \big\|_F^2 \\
                                                                 & &  \ \ \ \ \ \ \  +  \ \ \    \big\|\W_{\ell,\r}^{\tau} \big\|_F^2 .  \big\| \phii_{\ell-1}^\tau({\XX}_r)-  \phii_{\ell-1}^{\tau-1}({\XX}_r) \big\|_F^2   \ \ \ \ \ \ \ \ \ \ \ \ (\textrm{Cauchy Schwarz})\\
                                                                              & & \\   
                                                                              &=&     \big\|(\alphaa_{\ell,t}^{\tau})^\top  \   \P^\top \  \phii_{\ell-1}^\tau({\XX}_r) - ({\alphaa}_{\ell,t}^{\tau-1})^\top \  \P^\top \   \phii_{\ell-1}^{\tau-1}({\XX}_r) \big\|_F^2    \ \ \ \ \ \ \ \     (\textrm{Eq.}~\ref{eq1}~\textrm{in paper}) \\                         & &  \ \ \ \ \ \ \  +  \ \ \    \big\|\W_{\ell,\r}^{\tau} \big\|_F^2 .  \big\| \phii_{\ell-1}^\tau({\XX}_r)-  \phii_{\ell-1}^{\tau-1}({\XX}_r) \big\|_F^2 \\
                                                                              & & \\
                                                                              &=&     \big\| (\alphaa_{\ell,t}^{\tau})^\top \  \P^\top \P \ {\beta}_{\ell-1,r}^{\tau} - ({\alphaa}_{\ell,t}^{\tau-1})^\top \ \P^\top   \P \ {\beta}_{\ell-1,r}^{\tau-1}  \big\|_F^2,   \ \    \big(\phii_{\ell-1}^\tau({\XX}_r) =\P \ \beta_{\ell-1,r}^{\tau} \big) \\
                                                                              & &  \ \ \ \ \ \ \  +  \ \ \    \big\|\W_{\ell,\r}^{\tau} \big\|_F^2 .  \big\| \phii_{\ell-1}^\tau({\XX}_r)-  \phii_{\ell-1}^{\tau-1}({\XX}_r) \big\|_F^2 \\    
     &  &  \\ 
                                                                 &=&     \big\| (\alphaa_{\ell,t}^{\tau})^\top \ {\beta}_{\ell-1,r}^{\tau} - ({\alphaa}_{\ell,t}^{\tau-1})^\top  \ {\beta}_{\ell-1,r}^{\tau-1}  \big\|_F^2  \ \ \ \ \  \ \ \ \ \ \ \ \ \ \  \ \ \  \ \ \ \ \ \    ( \{\P_d\}_d \ \textrm{orthonormal}) \\
                                                                 & &  \ \ \ \ \ \ \  +  \ \ \    \big\|\W_{\ell,\r}^{\tau} \big\|_F^2 .  \big\| \phii_{\ell-1}^\tau({\XX}_r)-  \phii_{\ell-1}^{\tau-1}({\XX}_r) \big\|_F^2 \\
    & & \\
                                                   &\leq &    \big\| \alpha_{\ell,t}^\tau\big\|_F^2 . \big\| \beta_{\ell-1,r}^{\tau}   \big\|_F^2 +  \big\| {\alpha}_{\ell,t}^{\tau-1}\big\|_F^2 . \big\| {\beta}_{\ell-1,r}^{\tau-1} \big\|_F^2    \ \ \ \ \ \ \   \ \ \ \ \ \ \ \ \ \ \ \   (\textrm{Cauchy Schwarz})\\
        & &  \ \ \ \ \ \ \  +  \ \ \   \big\|\W_{\ell,\r}^{\tau} \big\|_F^2 .  \big\| \phii_{\ell-1}^\tau({\XX}_r)-  \phii_{\ell-1}^{\tau-1}({\XX}_r) \big\|_F^2. 
  \end{eqnarray*}
  Combining the above inequality using recursion  
  \begin{eqnarray*}\label{eq8}
    \big\|\phii_\ell^\tau({\XX}_r) - \phii_\ell^{\tau-1}({\XX}_r)  \big\|_F^2 & \leq &  \sum_{k=0}^{\ell-1}   \big(\big\| \alpha_{\ell-k,t}^\tau\big\|_F^2 . \big\| \beta_{\ell-k-1,r}^{\tau}   \big\|_F^2 +  \big\| {\alpha}_{\ell-k,t}^{\tau-1}\big\|_F^2 . \big\| {\beta}_{\ell-k-1,r}^{\tau-1} \big\|_F^2\big) .  \prod_{k'=0}^{k-1}  \big\|\W_{\ell-k',\r}^{\tau} \big\|_F^2\\
       &  &  \ \ \ \ \ \ \ \  + \ \prod_{k'=0}^{\ell-1}  \big\|\W_{\ell-k',\r}^{\tau} \big\|_F^2  . \big\|\phii_0^\tau({\XX}_r) - \phii_0^{\tau-1}({\XX}_r) \big\|_F^2 \\
                  & =&  \sum_{k=0}^{\ell-1}   \big(\big\| \alpha_{\ell-k,t}^\tau\big\|_F^2 . \big\| \beta_{\ell-k-1,r}^{\tau}   \big\|_F^2 +  \big\| {\alpha}_{\ell-k,t}^{\tau-1}\big\|_F^2 . \big\| {\beta}_{\ell-k-1,r}^{\tau-1} \big\|_F^2\big) .  \prod_{k'=0}^{k-1}  \big\|\W_{\ell-k',\r}^{\tau} \big\|_F^2,
   \end{eqnarray*}
    $\big\|\phii_0^\tau({\XX}_r) - \phii_0^{\tau-1}({\XX}_r)  \big\|_F^2  = \big\|{\XX}_r - {\XX}_{r} \big\|_F^2 =0$ as the parameters of the convolutional layers of the whole network $f$ are initially pretrained and fixed, so any incremental training of $f$ maintains $\{\phii_0({\XX}_r)\}_{r}$ unchanged. Considering $\eta$ the max number of epochs when training the parameters of the $t^\textrm{th}$ task, one may write

\begin{eqnarray*}\label{eq11}
   \big\|\phii_\ell^{\eta}({\XX}_r) - \phii_\ell^{0}({\XX}_r)  \big\|_F^2   &=& \big\|\phii_\ell^{\eta}({\XX}_r)  -  \sum_{\tau=1}^{\eta-1} \phii_\ell^\tau({\XX}_r)  +  \sum_{\tau=1}^{\eta-1} \phii_\ell^\tau({\XX}_r) -   \phii_\ell^{0}({\XX}_r)  \big\|_F^2   \\ 
   & \leq & \sum_{\tau=1}^{\eta}  \big\|\phii_\ell^\tau({\XX}_r) - \phii_\ell^{\tau-1}({\XX}_r) \big\|_F^2. 
\end{eqnarray*}
Combining the two above inequalities, it follows
  \begin{eqnarray*}
    \big\|\phii_\ell^{\eta}({\XX}_r)- \phii_\ell^0({\XX_r})  \big\|_F^2  &\leq&  \sum_{\tau=1}^{\eta}  \sum_{k=0}^{\ell-1}   \big(\big\| \alpha_{\ell-k,t}^\tau\big\|_F^2 . \big\| \beta_{\ell-k-1,r}^{\tau}   \big\|_F^2 +  \big\| {\alpha}_{\ell-k,t}^{\tau-1}\big\|_F^2  . \big\| {\beta}_{\ell-k-1,r}^{\tau-1} \big\|_F^2 \big) \\
   & &  \ \ \ \ \ \ \ \ \ \ \ \  . \prod_{k'=0}^{k-1}  \big\|\W_{\ell-k',\r}^{\tau} \big\|_F^2.
    \end{eqnarray*} 

\begin{flushright}
$\blacksquare$
\end{flushright}
}

\end{document}